\documentclass[10pt,twocolumn,letterpaper]{article}

\usepackage[pagenumbers]{cvpr} % To force page numbers, e.g. for an arXiv version

\usepackage{booktabs}       % professional-quality tables
\usepackage{amsfonts}       % blackboard math symbols
\usepackage{nicefrac}       % compact symbols for 1/2, etc.
\usepackage{microtype}      % microtypography
\usepackage{xcolor}         % colors
\usepackage{graphicx}
\usepackage{amsmath}
\usepackage{amssymb}
\usepackage{array}
\usepackage{multirow}
\usepackage{color}
\usepackage{colortbl}
\usepackage{framed}
\usepackage{bm}
\usepackage{xspace}
\usepackage{enumitem}
\usepackage{xparse}
\usepackage{algorithm}
\usepackage{listings}
\usepackage{url}
\usepackage{mathtools}
\usepackage{pifont}
\usepackage{mathtools}
\usepackage{lipsum}
\usepackage{adjustbox}
\usepackage[most]{tcolorbox}

\definecolor{lightgray}{gray}{0.92}
\definecolor{TitleColor}{gray}{0.95}

\definecolor{Gray}{gray}{0.93}
\definecolor{blond}{rgb}{0.98, 0.94, 0.75}
\definecolor{LightCyan}{rgb}{0.88,0.95,1}
\definecolor{OurColor}{HTML}{F9ECF3}

\def \ie {\emph{i.e.}}
\def \eg {\emph{e.g.}}

\newcommand{\tit}[1]{\smallbreak\noindent\textbf{#1.}}
\newcommand{\tinytit}[1]{\noindent\textbf{#1.}}

\newcommand{\ours}{VHS\xspace}

\definecolor{promptbg}{gray}{0.96} % Light gray background
\definecolor{promptframe}{gray}{0.70} % Darker gray frame

\lstdefinelanguage{LLMPrompt}{
    basicstyle=\ttfamily\small, % Typewriter font, small size
    breaklines=true,            % Wrap long lines automatically
    backgroundcolor=\color{promptbg},
    frame=single,               % Frame around the code
    rulecolor=\color{promptframe},
    frameround=ffff,            % Sharp corners (change to tttt for rounded)
    framesep=5pt,               % Padding between text and frame
    literate=
        {→}{$\rightarrow$}1
        {—}{---}1
        {“}{``}1
        {”}{''}1
        {…}{...}1
}

\definecolor{promptboxbg}{HTML}{F2F2F2}
\newtcolorbox{promptbox}[1][]{
  enhanced,
  breakable,
  colback=promptboxbg,
  colframe=promptboxbg!20!black,
  leftrule=1.5mm,
  arc=1mm,
  boxrule=0.6pt,
  top=2mm, bottom=2mm, left=3mm, right=3mm,
  fonttitle=\bfseries,
  title=Prompt,
  fontupper=\ttfamily,
  #1,
  before upper={\color{black}},
}

\definecolor{promptboxbg}{HTML}{F2F2F2}

\lstdefinelanguage{LLMPrompt}{
    basicstyle=\ttfamily\footnotesize,
    breaklines=true,
    backgroundcolor={},    % no background
    frame=none,            % no border
    frameround=ffff,
    breakindent=0pt,        % continuation lines start at column 0
    breakautoindent=true, 
    xleftmargin=10pt,       % <-- NO INDENT
    xrightmargin=10pt,      % <-- NO INDENT
    aboveskip=5pt,
    belowskip=5pt,
    literate=
        {→}{$\rightarrow$}1
        {—}{---}1
        {“}{``}1
        {”}{''}1
        {…}{...}1
}

\newtcblisting{prompt}[1][]{
    enhanced,
    listing only,
    listing options={
        language=LLMPrompt,
        basicstyle=\ttfamily\footnotesize,
        xleftmargin=10pt,
        xrightmargin=10pt,
    },
    colback=promptboxbg,
    colframe=promptboxbg!20!black,
    leftrule=1.5mm,
    arc=1mm,
    boxrule=1pt,
    top=1mm, bottom=1mm, left=1mm, right=1mm,
    fonttitle=\bfseries,
    title=#1,
    titlerule=0pt,              
}

\definecolor{cvprblue}{rgb}{0.21,0.49,0.74}
\usepackage[pagebackref,breaklinks,colorlinks,allcolors=cvprblue]{hyperref}

\def\paperID{****} % *** Enter the Paper ID here
\def\confName{CVPR}
\def\confYear{2026}

\title{Tiny Inference-Time Scaling with Latent Verifiers}

\author{Davide Bucciarelli$^{*1,2}$ \quad Evelyn Turri$^{*1}$ \quad Lorenzo Baraldi$^{2}$, \\ 
Marcella Cornia$^{1}$ \quad Lorenzo Baraldi$^{1}$ \quad Rita Cucchiara$^{1}$\\ 
$^1$University of Modena and Reggio Emilia, Italy \quad $^2$University of Pisa, Italy \\ 
{\tt\small $^1$\{name.surname\}@unimore.it, $^2$\{name.surname\}@phd.unipi.it} \\
{\tt\small \href{https://aimagelab.github.io/VHS}{aimagelab.github.io/VHS} }
}

\begin{document}
\maketitle
\begin{abstract}
Inference-time scaling has emerged as an effective way to improve generative models at test time by using a verifier to score and select candidate outputs. A common choice is to employ Multimodal Large Language Models (MLLMs) as verifiers, which can improve performance but introduce substantial inference-time cost. Indeed, diffusion pipelines operate in an autoencoder latent space to reduce computation, yet MLLM verifiers still require decoding candidates to pixel space and re-encoding them into the visual embedding space, leading to redundant and costly operations. In this work, we propose \underline{\textbf{V}}erifier on \underline{\textbf{H}}idden \underline{\textbf{S}}tates (\textbf{\ours}), a verifier that operates directly on intermediate hidden representations of Diffusion Transformer (DiT) single-step generators. \ours analyzes generator features without decoding to pixel space, thereby reducing the per-candidate verification cost while improving or matching the performance of MLLM-based competitors. We show that, under tiny inference budgets with only a small number of candidates per prompt, \ours enables more efficient inference-time scaling reducing joint generation-and-verification time by 63.3\%, compute FLOPs by 51\% and VRAM usage by 14.5\% with respect to a standard MLLM verifier, achieving a +2.7\% improvement on GenEval at the same inference-time budget.
\end{abstract}    
\section{Introduction}
\label{sec:intro}
\begin{figure}[t]
    \centering
    \includegraphics[width=0.98\linewidth]{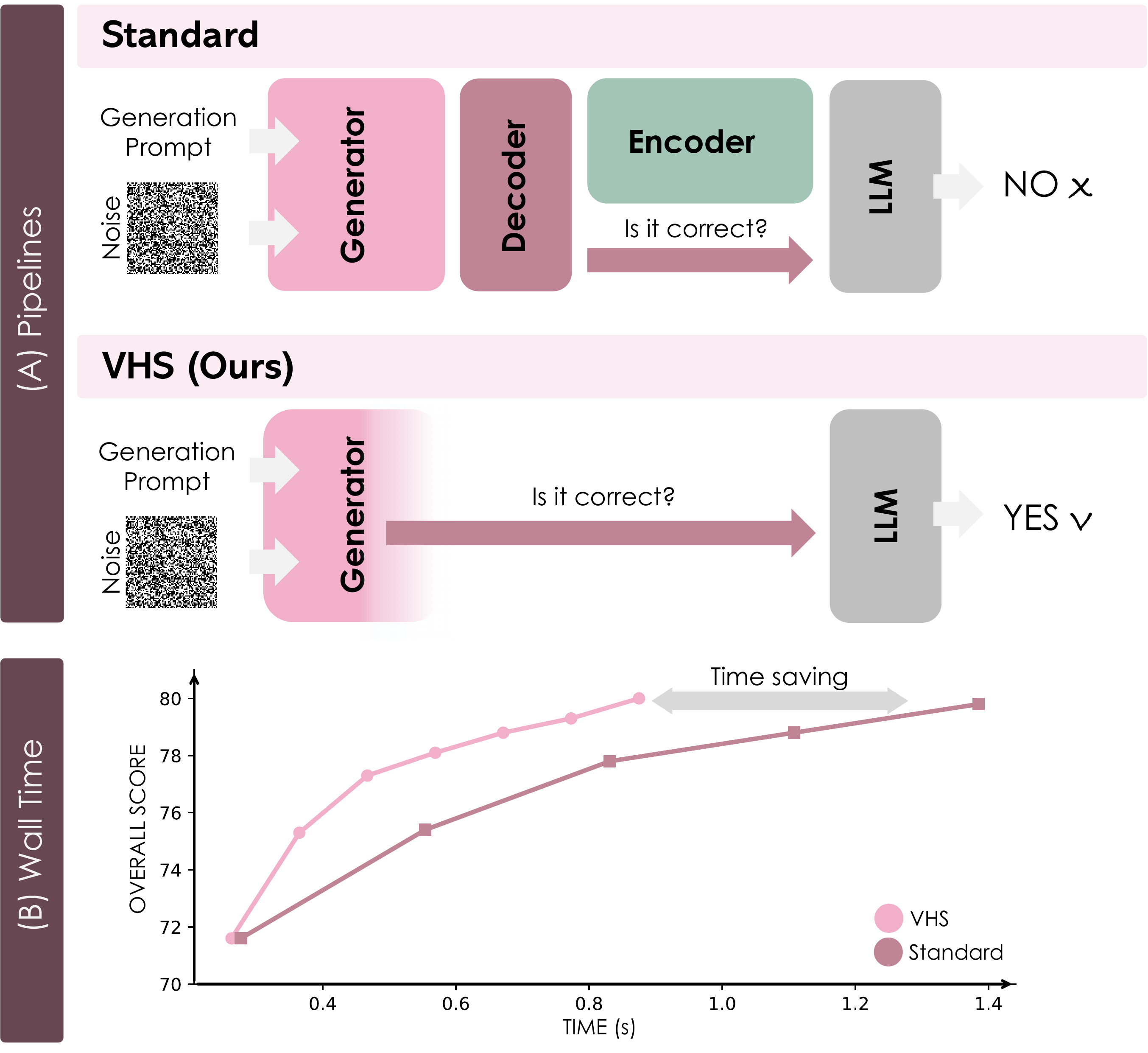}
    \caption{(A) Comparison between standard inference-time scaling and \ours. \ours skips part of the generation pipeline and avoids the decoding and re-encoding steps. (B) \ours achieves a comparable quality score on GenEval~\cite{ghosh2023geneval} in just 57\% of the compute time.}
    \vspace{-0.45cm}
    \label{fig:teaser}
\end{figure}
Diffusion and flow-based models~\cite{ho2020denoising, song2020denoising, lipman2022flow} have recently transformed image synthesis, producing samples that closely resemble natural imagery with remarkable fidelity and controllability. However, their generation process remains computationally expensive and often misaligned with user intent. To mitigate these limitations, recent works have adopted the inference-time scaling paradigm~\cite{xie2025sana,kim2025inference,singh2025code,ma2025inference}, which allocates additional computational budget at inference by generating multiple candidate samples and selecting the most suited among them. This framework relies on two key components: (i) an exploration algorithm that generates multiple candidates, and (ii) a verifier that assigns scores to the candidates and selects those that best match the prompt.

In image generation, verifiers are typically implemented by fine-tuning Multimodal Large Language Models (MLLMs)~\cite{caffagni2024revolution} with an image scoring objective. Nevertheless, MLLMs are computationally heavy, and their inference cost is not negligible. Despite this, recent literature~\cite{ma2025inference,kim2025inference} mostly accounts for the number of function evaluations (\eg, diffusion steps), while treating the cost of the verifier as implicit overhead, leading to an incomplete view of the computational footprint of inference-time scaling. Moreover, many existing evaluations assume very large budgets, sometimes involving thousands of function evaluations~\cite{kim2025inference}, whereas practical deployment scenarios, such as commercial image generation services, typically operate under much tighter constraints, often returning only a handful of images (\eg, up to four candidates) per prompt. Further, visual generative models typically operate in a compressed latent space~\cite{rombach2022high}, defined by an autoencoder, whereas MLLMs rely on an external visual encoder (\eg, CLIP~\cite{radford2021learning}) to obtain image representations. Thus, to score a generated sample, the latent representation must be decoded into pixel space and then re-encoded by the visual backbone of the MLLM.
Although this decode-encode overhead may be acceptable in standard multi-step generators, it becomes increasingly significant in the case of single-step image generators, which can produce images in a single function evaluation~\cite{chen2025sana,Yin_2024_CVPR,sauer2024adversarial,luo2023latent}.

Following these considerations, we argue that the architecture of MLLM-based verifiers~\cite{he-etal-2024-videoscore,xie2025sana} should be reconsidered in light of the specific characteristics of the task. To this aim, we introduce \underline{\textbf{V}}erifier on \underline{\textbf{H}}idden \underline{\textbf{S}}tates (\textbf{\ours}), a MLLM-based verifier that directly aligns internal hidden representations of image generators with the embedding space of an LLM. Concretely, \ours operates on single-step image generators, extracting latent features during the generation process, and uses these hidden states as the visual inputs to the LLM (Fig.~\ref{fig:teaser}). This way, \ours eliminates the encoding-decoding overhead in the evaluation step, enabling significantly more efficient verification within the inference-time scaling framework, while retaining the expressivity of MLLM-based scoring. As a consequence, \ours is well-suited for tiny computational budgets, where only a small number of candidates per prompt is affordable, and thus closely aligns with the practical constraints and deployment settings of real-world commercial image generation services.

We evaluate \ours in terms of latency and verification quality on the GenEval benchmark~\cite{ghosh2023geneval}. In combination with the single-step generator SANA-Sprint~\cite{chen2025sana} and a compact LLM (Qwen2.5-0.5B~\cite{bai2025qwen2}), \ours reduces the joint generation-and-verification time by 63.3\% of that required by a standard MLLM-based verifier. Furthermore, under matched wall-clock budgets, \ours improves inference-time scaling performance on GenEval, achieving overall score gains of 3.1\%, 1.7\%, and 0.5\% over a CLIP-based MLLM verifier in the Best-of-2, Best-of-4, and Best-of-6 settings, respectively.

\noindent In summary, our main contributions are:
\begin{itemize}
    \item We introduce \ours, a verifier that operates directly on internal hidden states of DiT-based image generators, aligning visual latents with an LLM without passing through pixel space or an external visual encoder.
    \item We define a latency-aware inference-time scaling setting for single-step image generation, explicitly measuring wall-clock time and analyzing performance in realistic few-sample generation regimes.
    \item We provide a thorough empirical study of verifier design and latency, comparing alternative architectures and configurations (\ie, in terms of layers, backbones, and loss functions) and quantifying the trade-offs between computational cost and semantic alignment.
\end{itemize}

\section{Related Work}
\label{sec:related}
\tinytit{Image Generation Techniques}
Image generation has advanced substantially with the advent of diffusion models~\cite{ho2020denoising,song2020denoising}, which have surpassed GANs~\cite{goodfellow2014generative} in both sample quality and training stability. Latent diffusion models~\cite{rombach2022high} further extended this progress by operating in a compressed latent space, enabling high-resolution synthesis at a manageable computational cost. While early diffusion architectures relied primarily on U-Nets~\cite{ronneberger2015u} for noise prediction, these have recently been surpassed by Diffusion Transformers (DiTs)~\cite{Peebles_2023_ICCV}, which offer improved scalability and performance. In parallel, flow-based approaches~\cite{lipman2022flow,esser2024scaling} have reformulated the diffusion objective from noise estimation to velocity field prediction, providing an alternative yet closely related view of the generative process.

A complementary line of research focuses on improving inference efficiency. Few-step and even single-step diffusion models have been developed via distillation, making it possible to generate high-fidelity images with only a handful (or even a single) denoising step~\cite{luo2023latent,chen2025sana,sauer2024adversarial,Yin_2024_CVPR}. In this area, Stable Diffusion XL-Turbo~\cite{sauer2024adversarial} introduced adversarial diffusion distillation to ensure high-fidelity synthesis in the low-step regime and leveraged large pre-trained multi-step models as teachers, with a mixture of adversarial training and score distillation. Subsequently, PixArt-$\alpha$-DMD~\cite{Yin_2024_CVPR} proposed a distribution-matching distillation approach to align the student with the teacher model at the distribution level. Differently, SANA-Sprint~\cite{chen2025sana} presented a hybrid distillation framework that combines training-free continuous-time consistency distillation with latent adversarial distillation, and enables efficient adaptation of pre-trained diffusion or flow-matching models in the few-step generation scenario. 

\tit{Inference-Time Scaling}
Inference-time scaling~\cite{snell2024scaling} consists in allocating additional computational resources during inference to improve model performance, rather than increasing compute during training. This strategy, widely adopted in NLP for LLM inference~\cite{guan2025rstar,snell2024scaling,wu2025inference}, has been recently extended to visual content generation~\cite{xie2025sana,kim2025inference,singh2025code,ma2025inference,baraldi2025verifier}. In this context, it has been shown that allocating more compute time, beyond simply increasing the number of diffusion steps~\cite{ma2025inference}, can significantly enhance generation quality. 

Inference-time scaling methods for visual generation typically rely on two main components: a search algorithm and a verifier. The former generates candidate samples, while the latter evaluates and ranks them to select the best output. The simplest strategy, Best-of-\textit{N}, independently samples and scores \textit{N} candidates, selecting the highest-scoring one as the final result. Another algorithm, widely adopted in LLM inference-time scaling is beam search~\cite{snell2024scaling}, a heuristic algorithm that maintains the top-$k$ most probable candidates at each step, balancing exploration and efficiency to improve generation quality over greedy sampling. 

On the other end, verifiers are often based on MLLMs, leveraging their ability to interpret complex prompts and assess visual-textual alignment in the generated content. For instance, VQA-Score~\cite{lin2024evaluating} employs a Visual Question Answering model that scores samples based on the probability of the ``yes'' token in response to predefined questions assessing prompt fulfillment. Similarly, Vision-Reward~\cite{xu2024visionreward} queries an MLLM with fine-grained binary questions and combines the results through a learned weighting scheme.

In contrast with previous literature, we propose a verifier that directly works in the latent space of the generator, significantly reducing the computational overhead of verification.

\tit{Multimodal Large Language Models}
MLLMs extend traditional language models by integrating information across multiple modalities~\cite{caffagni2024revolution,liu2023visual,liu2024improved,tong2024cambrian,lin2024vila}, most notably, vision and text. Common architectures rely on a pretrained image encoder~\cite{radford2021learning,cherti2023reproducible,zhai2023sigmoid} whose embeddings are projected into the input space of the LLM through a lightweight adapter. This design allows the visual features to be integrated seamlessly into the token sequence of the LLM, enabling multimodal understanding and grounded generation. 

This framework was popularized by LLaVA~\cite{liu2023visual,liu2024improved}, which employed simple linear layers as connector and introduced a two-stage training pipeline: aligning the connector using image-caption pairs, and subsequently fine-tuning the entire model on instruction-following datasets. Building on this, several works have proposed to improve visual grounding and fine-grained alignment. Idefics3~\cite{laurenccon2024building} partitions images into spatial tiles encoded independently, improving localization and detailed perception. Similarly, Qwen2.5-VL~\cite{bai2025qwen2} incorporates 2D positional encodings into token representations to better preserve spatial structure within images. In contrast, we directly align hidden states of a DiT-based generator with the LLM, enabling image evaluation from latent representations rather than decoded pixels.

\section{Proposed Method}
\label{sec:method}
\begin{figure*}[t]
    \centering
    \includegraphics[width=\linewidth]{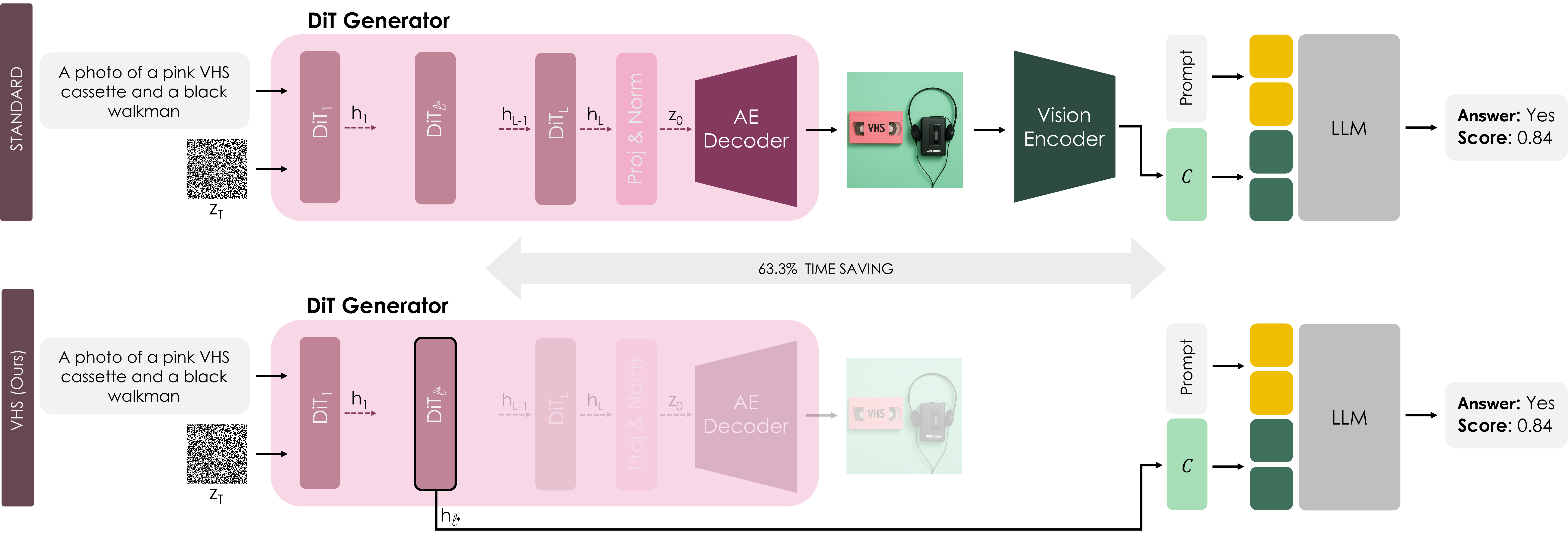}
    \vspace{-0.25cm}
    \caption{Comparison between a standard generation-verification pipeline (top) and \ours (bottom). \ours consumes visual features directly from the hidden states of the generator, bypassing subsequent DiT layers, autoencoder (AE) decoding, and CLIP-based re-encoding, significantly reducing sampling and verification overhead.}
    \vspace{-0.35cm}
    \label{fig:method}
\end{figure*}
\subsection{Preliminaries}
The objective of our approach is to assess content quality directly from the latent representations of a single-step image generator. In the following, we first formalize the generative process of multi-step models and subsequently introduce the single-step formulation adopted in our method.

Visual generative models, such as diffusion~\cite{ho2020denoising,rombach2022high} and flow-based models~\cite{lipman2022flow,esser2024scaling,xie2025sana}, synthesize data through a multi-step refinement process. Starting from a latent variable sampled from a prior distribution, $z_T \sim p_T$ (\eg~$\mathcal{N}(0, \mathbf{I})$), the model progressively refines it into a structured representation $z_0$ by constructing a discrete trajectory
\begin{equation}
z_T \rightarrow z_{T-1} \rightarrow \dots \rightarrow z_1 \rightarrow z_0,
\end{equation}
where transition steps are parameterized by a neural network $f_\theta$ that predicts model-specific quantities such as noise or velocity, depending on the underlying framework. This iterative process transports samples from the prior $p_T$ toward an approximation of the target distribution $p_{\text{data}}$, yielding a sequence $\{z_t\}_{t=T}^0$ that we refer to as the \emph{generative trajectory}.

Concretely, in DiT-based~\cite{Peebles_2023_ICCV,esser2024scaling} generators, the noisy latent $z_t$ is processed by a Transformer backbone that produces a sequence of hidden representations $\{h_\ell\}_{\ell=0}^{L-1}$ generated with $h_\ell = \mathrm{DiT}_\ell(h_{\ell-1}, t)$, where $L$ is the number of layers in the DiT and $h_0$ is the noisy latent $z_t$. Lastly, the DiT layers are followed by a decoder  $\mathcal{D}$ that operates on the final hidden state $h_{L-1}$ after projection and normalization (\ie, $z_0$). 
In both diffusion~\cite{rombach2022high} and flow models, the generation trajectory is defined not in pixel space but in the compressed latent space of an autoencoder~\cite{chen2024deep}, which we define as $\mathcal{E}$. During sampling, the generative trajectory $\{z_t\}_{t=0}^T$ evolves entirely in $\mathcal{E}$, and the final image is obtained by decoding the terminal latent $z_0$ via
$x_0 = \mathcal{D}(z_0)$.

In contrast, a single-step generator is obtained by distilling a standard diffusion or flow-based model into a network that maps a latent sample $z_T \sim \mathcal{N}(0, I)$ to an image in one forward pass producing a generative trajectory with $T=1$. 
While in multi-step diffusion and flow-based models the computational cost of the decoding operator $\mathcal{D}$ is typically negligible compared to the iterative sampling process, in single-step generators~\cite{chen2025sana,Yin_2024_CVPR,sauer2024adversarial} this balance shifts: the forward pass of $\mathcal{D}$ becomes a non-trivial component of the total inference cost. For this reason, our method operates directly on the intermediate latent representation $h$ when tasked with verifying generated samples, thereby skipping the forward pass through $\mathcal{D}$ and avoiding any decoding overhead.

\subsection{Latent Verifier}
Within the inference-time scaling framework, a key component is the verifier model, which evaluates generated samples and identifies the most promising ones. In our formulation, we define the verifier as a model $\mathcal{S}_\theta$ that, given a generated sample $x_0$ and the user prompt $p$, outputs $s \in \{\text{Yes}, \text{No}\}$ indicating whether the sample is semantically aligned with~$p$. Recent works~\cite{xie2025sana,li2025reflect,kim2025inference,zhuo2025reflection} typically implement verifiers using MLLMs. Although such models have shown strong performance in assessing generation quality, the computational cost associated with their scoring procedure is non-negligible. Nevertheless, most inference-time scaling studies~\cite{ma2025inference,kim2025inference} quantify the computational budget solely by the number of generator function evaluations (\eg, diffusion steps), with the cost of running the MLLM-based verifier either implicitly ignored or treated as negligible.

Formally, an MLLM-based verifier can be decomposed into three components: (i) a visual encoder $\mathcal{V}$, which maps an input image $x_0$ to a sequence of visual tokens; (ii) a connector $\mathcal{C}$, which projects these visual tokens into the embedding space of the language model; and (iii) a language model, which performs multimodal reasoning over the concatenated visual and textual tokens and produces the final score. In the inference-time scaling setting, this architecture is used as follows: a latent sample $z_0$ is drawn from the latent space of the generator, decoded into pixel space as $x_0 = \mathcal{D}(z_0)$, and then processed by the verifier to produce a score
\begin{equation}
\label{eq:mllm}
    s = \mathcal{S}_\theta(z_0, p)
      = \mathrm{LLM}\big(\mathcal{C}(\mathcal{V}(\mathcal{D}(z_0))), p\big).
\end{equation}

In this pipeline, $\mathcal{V}$ is responsible only for re-encoding visual information that has already been implicitly represented in the latent space of the generator~\cite{Xu_2023_CVPR,NEURIPS2023_0503f5dc,PNVR_2023_ICCV}. We claim that for generative models that operate in a rich latent space, this additional pass through $\mathcal{V}$ is not semantically essential for the verification task. Instead, it does introduce a non-trivial decode-encode overhead: the latent $z_0$ must undergo two successive transformations, $\mathcal{D}(z_0)$ and $\mathcal{V}(\cdot)$, before the LLM can reason about the sample.

Our \underline{\textbf{V}}erifier on \underline{\textbf{H}}idden \underline{\textbf{S}}tates (\textbf{\ours}) explicitly bypasses the decoding-encoding bottleneck by removing the visual encoder from the verification loop. Instead of operating on the decoded image, \ours directly consumes hidden representations from the generator. Specifically, \ours acts on the output of a DiT layer $\ell^\star \in \{0,\dots,{L-1}\}$, denoted as $h_{\ell^\star}$, and feeds it to the connector $\mathcal{C}$ of the MLLM, as follows:
\begin{equation}
\label{eq:vhs}
    s = \mathcal{S}_\theta(z_0, p)
      = \mathrm{LLM}\big(\mathcal{C}(h_{\ell^\star}), p\big),
\end{equation}
where the connector $\mathcal{C}$ is trained to align $h_{\ell^\star}$ with the LLM input space, treating hidden features like image features.

This design yields two key advantages. First, it completely removes the decoding-encoding pipeline
$z_0 \rightarrow x_0 \rightarrow \mathcal{V}(x_0)$ from the verification process, thereby reducing per-sample evaluation latency. Second, since \ours accesses hidden states at layer $\ell^\star$, it allows us to truncate the generator during verification and skip the remaining $L - (\ell^\star + 1)$ layers. As a result, \ours provides semantically informed verification at a fraction of the computational cost of standard MLLM-based verifiers, making inference-time scaling substantially more practical in low-latency generation regimes.  An overview of our approach is shown in Fig.~\ref{fig:method}, in comparison with a standard generation-verification pipeline.

\subsection{Training Procedure Overview}
\label{sec:training}
\ours is trained via a two-stage procedure. First, in an alignment stage, we adapt the visual representation from the generator hidden layers to be compatible with the LLM backbone. Second, we fine-tune the model as a~verifier.

\tit{Alignment Stage}
In this stage, the goal is to align the visual representations extracted from $h_{\ell^\star}$ with the representation space of the LLM. Unlike standard visual encoders, our visual embedder is a generative model. As a consequence, we first need to generate raw images to obtain the intermediate features $h_{\ell^\star}$ used for alignment. Concretely, we build upon the dataset used in the first stage of the LLaVA training~\cite{liu2023visual}, which provides image-caption pairs usually employed to train MLLMs. Starting from each caption, we employ the generator to produce a synthetic image and record the associated hidden representation $h_{\ell^\star}$. Notably, this may introduce inconsistencies between the original caption and the generated image, due to hallucinations or semantic drift in the generator. To mitigate this, we re-caption each generated image using Gemma-3-4B~\cite{team2025gemma}, and use the resulting captions as the textual supervision for the alignment stage.

\begin{table*}[t]
  \centering
  \setlength{\tabcolsep}{0.6em}
  \caption{End-to-end inference time, FLOPs and VRAM usage for Best-of-$N$ generation with SANA-Sprint~\cite{chen2025sana} under different computational budgets, along with the relative savings (\%) compared to the standard verifier.}
  \vspace{-0.15cm}
  \resizebox{\linewidth}{!}{
  \begin{tabular}{l c c c c c c c c c c c c c c c c c c}
    \toprule
     & & \multicolumn{5}{c}{\textbf{Inference Time (ms)}} & & \multicolumn{5}{c}{\textbf{TFLOPs}} & & \multicolumn{5}{c}{\textbf{Peak VRAM Usage (GB)}} \\
     \cmidrule(lr){3-7} \cmidrule(lr){9-13} \cmidrule(lr){15-19}
     & & \textbf{Saved (\%)} & $\mathsf{Bo1}$ &  $\mathsf{Bo2}$ & $\mathsf{Bo4}$ & $\mathsf{Bo6}$ & & \textbf{Saved (\%)} & $\mathsf{Bo1}$ & $\mathsf{Bo2}$ & $\mathsf{Bo4}$ & $\mathsf{Bo6}$ & & \textbf{Saved (\%)} & $\mathsf{Bo1}$ & $\mathsf{Bo2}$ & $\mathsf{Bo4}$ & $\mathsf{Bo6}$ \\
    \midrule
    MLLM w/ CLIP & & - & 277 & 554 & 1108 & 1662 & & - & 15.1 & 28.5 & 55.1 & 81.8 & & - & 13.8 & 15.5 & 18.8 & 22.2\\
    MLLM w/ AE & & 50.2\% & 138 & 401 & 677 & 953 & & 51.0\% & 7.4 & 14.8 & 29.5 & 44.3 & & \textbf{14.5\%} & \textbf{11.8} & \textbf{11.9} & \textbf{12.3} & \textbf{12.6}\\
    \midrule
    \rowcolor{OurColor}
    \ours on $h_{7}$ & & \textbf{63.3\%} & \textbf{102} & \textbf{363} & \textbf{565} & \textbf{767} & & \textbf{62.9\%} & \textbf{5.6} & \textbf{11.3} & \textbf{22.5} & \textbf{33.8} & & \textbf{14.5\%} & \textbf{11.8} & \textbf{11.9} & \textbf{12.3} & \textbf{12.6}\\
    \bottomrule
  \end{tabular}
  }
  \label{tab:inference_time}
  \vspace{-0.2cm}
\end{table*}

\tit{Verifier Fine-tuning}
While the alignment stage is largely consistent with standard MLLM training, the verifier fine-tuning stage explicitly adapts the model to the scoring objective required in the inference-time scaling setting. Building on existing literature~\cite{li2025reflect}, we adopt the prompts of the training dataset of Reflect-DiT~\cite{li2025reflect} and generate 20 candidate images per prompt, resulting in a total of 118k samples.  These candidates are categorized by Gemma-3-4B~\cite{team2025gemma} into the respective GenEval categories~\cite{ghosh2023geneval} and scored with its automatic evaluator. Based on these evaluation scores, we derive binary labels (Yes/No) for each image in the dataset.

Analysis of the training set and GenEval benchmark reveals a significant class imbalance, with correct samples substantially overrepresented. A uniform weighting scheme in the training loss consequently underemphasizes the minority ``incorrect'' class, leading to suboptimal verifier performance. To address this, \ours employs a weighted cross-entropy loss during verifier fine-tuning. This approach re-weights the training signal proportionally to class frequencies, effectively compensating for the skewed label distribution and improving model calibration on underrepresented samples.
\section{Experimental Results}

\subsection{Implementation Details}
In the alignment stage, we follow the LLaVA~\cite{liu2023visual} training scheme, and tune only the newly initialized connector module during the first stage. Differently, in the verifier fine-tuning stage, we train both the connector and the whole language model, splitting the generated datasets in training (80\%) and evaluation (20\%), selecting the model yielding the best evaluation loss. All models follow the exact same training procedure, ensuring a fair comparison between models trained with equivalent data and policy. 

To derive a more granular scoring mechanism from binary labels, we leverage the LLM output probability of the sampled token (``yes'' or ``no'') to produce a continuous score. Best-of-N selection is then performed by retaining the highest-scoring sample according to this approach. We refer the reader to the supplementary materials for a detailed explanation of this approach and accompanying ablation~studies.

\begin{table*}[t]
  \centering
  \setlength{\tabcolsep}{0.8em}
  \caption{Accuracy (\%) on the GenEval benchmark~\cite{ghosh2023geneval} across computational budgets, generator backbones, and verifier configurations (on LLM Qwen2.5-0.5B). Results compare SANA-1.5 and SANA-Sprint~\cite{chen2025sana} under matched wall-clock budgets (milliseconds), with each verifier operating under the same time constraint via adaptive Best-of-$N$.}
  \vspace{-0.15cm}
  \resizebox{\linewidth}{!}{
  \begin{tabular}{c c c c l c c c c c c c}
    \toprule
     \textbf{Budget} & \textbf{Generator} & \textbf{Steps} & \textbf{Verifier} & \textbf{Best-of-N} &  \textbf{Single} & \textbf{Two} & \textbf{Counting} & \textbf{Color} & \textbf{Position} & \textbf{Attribution} & \textbf{Overall} \\
    \midrule
    200ms & SANA-Sprint & 1 & - & Best-of-1 & 99.3 & 88.1 & 56.0 & 87.6 & 54.1 & 47.8 & 71.6 \\
    \midrule
    \multirow{6}{*}{550ms}
    & SANA-1.5 & 4 & - & Best-of-1 & 98.8 & 78.2 & 66.5 & 71.1 & 50.6 & 20.8 & 63.0 \\
    & SANA-Sprint & 8 & - & Best-of-1 & 99.5 & 91.9 & 59.3 & 86.0 & 57.8 & 52.4 & 74.0 \\
    \cmidrule{2-12}
    & SANA-Sprint & 1 & MLLM w/ CLIP & Best-of-2 & \textbf{100.0} & 91.3 & 59.5 & 88.0 & 61.0 & 55.4 & 75.4 \\
    & SANA-Sprint & 1 & MLLM w/ AE & Best-of-3 & \textbf{100.0} & 90.9 & 59.0 & 89.6 & 55.8 & 50.6 & 73.1  \\
     \rowcolor{OurColor}
     & SANA-Sprint & 1 & \textbf{\ours (Ours)} & Best-of-4 & \textbf{100.0} & \textbf{93.9} & \textbf{61.5} & \textbf{90.6} & \textbf{66.2} & \textbf{58.4} & \textbf{78.1} \\
    \midrule
    \multirow{6}{*}{1100ms}
     & SANA-1.5 & 12 & - & Best-of-1 & 100.0 & 92.7 & 74.8 & 88.3 & 61.4 & 59.6 & 78.8 \\
    & SANA-Sprint & 20 & - & Best-of-1 & 100.0 & 88.5 & 59.8 & 89.6 & 48.6 & 51.0 & 72.2 \\
    \cmidrule{2-12}
    & SANA-Sprint & 1 & MLLM w/ CLIP & Best-of-4 & \textbf{100.0} & 92.7 & 66.0 & 88.9 & 65.9 & 61.6 & 78.8 \\
    & SANA-Sprint & 1 & MLLM w/ AE & Best-of-7 & 99.7 & 90.7 & 61.3 & \textbf{90.8} & 59.6 & 49.3 & 74.7 \\
     \rowcolor{OurColor}
    & SANA-Sprint & 1 & \textbf{\ours (Ours)} & Best-of-9 & \textbf{100.0} & \textbf{95.7} & \textbf{66.5} & 88.9 & \textbf{69.8} & \textbf{63.8} & \textbf{80.5} \\   
    \midrule
    \multirow{6}{*}{1650ms}
    & SANA-1.5 & 16 & - & Best-of-1 & 99.7 & 93.5 & 77.3 & 89.1 & 60.2 & 60.8 & 79.4\\
    & SANA-Sprint & 30 & - & Best-of-1 & 100.0 & 90.5 & 57.3 & 85.1 & 49.3 & 50.2 & 71.4 \\
    \cmidrule{2-12}
    & SANA-Sprint & 1 & MLLM w/ CLIP & Best-of-6 & \textbf{100.0} & 93.9 & \textbf{68.2} & 88.7 & 69.8 & 64.2 & 80.4 \\
    & SANA-Sprint & 1 & MLLM w/ AE & Best-of-11 & 99.7 & 90.5 & 59.3 & \textbf{89.8} & 58.4 & 49.0 & 73.9 \\
     \rowcolor{OurColor}
    & SANA-Sprint & 1 & \textbf{\ours (Ours)} & Best-of-15 & \textbf{100.0} & \textbf{96.0} & 67.3 & 89.1 & \textbf{70.4} & \textbf{64.6} & \textbf{80.9} \\   
    \bottomrule
  \end{tabular}
  }
  \label{tab:geneval}
  \vspace{-0.4cm}
\end{table*}

\subsection{Experimental Setting}
To ensure a fair evaluation of our approach, we adopt a controlled experimental setup and define three verifier architectures, namely:
(i) \textbf{MLLM w/ CLIP}, a standard MLLM following the LLaVA design~\cite{liu2023visual}, where a frozen CLIP encoder (using the ViT-L/14@336 variant) is connected to the LLM through an MLP projection layer. This configuration is the only one that employs an external visual encoder;
(ii) \textbf{MLLM w/ AE}, a variant in which the latent output of the generator, $z_0$, is mapped to the LLM input space via an MLP. This is equivalent to encoding the image with the autoencoder encoder $\mathcal{E}$ and processing the latent representation through the connector;
(iii) \textbf{\ours}, our proposed model, which feeds an intermediate hidden representation $h_\ell$ from the generator into the LLM using the same linear projection layer adopted in the other architectures.
To precisely quantify the time savings achieved by different evaluators, we report measurements averaged over 10 runs following an initial warm-up phase to stabilize performance. Specifically, we measure the time required for a full generation-and-evaluation cycle and the time required to generate up to $N$ images and select the best one. All experiments are conducted on the SANA-Sprint generator~\cite{chen2025sana} with a single step of generation, with an NVIDIA A100 GPU.

\subsection{Latency Estimation of \ours}
Table~\ref{tab:inference_time} reports inference costs across three axes: wall-clock time, FLOPs, and peak VRAM usage, for both baselines and \ours, evaluated under Best-of-$N$ selection with SANA-Sprint. Time savings are expressed as a percentage relative to MLLM w/ CLIP, which we regard as the standard verifier.

Across all dimensions, the results consistently show that bypassing the decoding–encoding operation ($\mathcal{V}(D(z_0))$) required by MLLM w/ CLIP yields substantial gains. Replacing the CLIP-based verifier with the AE-based one (MLLM w/ AE) already halves the cost: inference time drops from 277\,ms to 138\,ms ($-$50.2\%), FLOPs are reduced by 51.0\%, and peak VRAM consumption falls from 13.8\,GB to 11.8\,GB ($-$14.5\%). Skipping part of the DiT forward, \ours pushes these savings further: the best configuration, \ours on $h_7$, reaches 102\,ms ($-$63.3\%), reduces FLOPs by 62.9\%, matching the VRAM footprint of MLLM w/ AE.

These gains translate directly into end-to-end efficiency under Best-of-$N$ selection. In the $\mathsf{Bo6}$ setting, MLLM w/ CLIP requires 1662\,ms, MLLM w/ AE reduces this to 953\,ms, and \ours on $h_7$ further decreases it to 767\,ms. Crucially, under a time budget comparable to MLLM w/ CLIP at $\mathsf{Bo3}$ (831\,ms), \ours on $h_7$ can already afford $\mathsf{Bo6}$, effectively doubling the candidate pool. The computational savings across time, FLOPs, and memory are thus not merely theoretical: they can be directly traded for a larger candidate pool under the same wall-clock and hardware budget.

\subsection{Performance on GenEval}
Beyond comparing the latency of \ours against the MLLM w/ CLIP and MLLM w/ AE baselines, we now turn to evaluating the verifier performance in an inference-time scaling benchmark, where the available computational budget is explicitly constrained in terms of wall-clock time.

\tit{Experimental Setup}
We conduct this analysis on the GenEval benchmark~\cite{ghosh2023geneval}, which evaluates generator performance across six categories: Single Object, Two Objects, Counting, Colors, Position, and Attribute Binding. Each category is defined by structured prompts designed to probe specific capabilities of the generator. For instance, ``Counting'' requires producing a specific number of objects, while ``Position'' involves rendering two objects in a fixed spatial configuration. We conduct comparisons across three different time settings (550\,ms, 1100\,ms, and 1650\,ms), which approximately correspond to the wall-clock time required by the MLLM w/ CLIP to produce the best of 2, 4, and 6 generations, respectively. In contrast, for the MLLM variants using AE and VHS, the time savings (cfr. Table~\ref{tab:inference_time}) allow us to perform a wider sample exploration within the same computational budget. Specifically, 3 and 4 generations in 550\,ms, 7 and 9 in 1100\,ms, and 11 and 15 in 1650\,ms, respectively.

\begin{table*}[t]
  \centering
  \setlength{\tabcolsep}{1.em}
  \caption{Accuracy (\%) of SANA-Sprint~\cite{chen2025sana} on the GenEval benchmark~\cite{ghosh2023geneval} across varying hidden layers, training losses, training data, and reference LLMs on a time budget of 1100\,ms.}
  \vspace{-0.15cm}
  \resizebox{\linewidth}{!}{
  \begin{tabular}{l c c c c c c c}
    \toprule
     \textbf{Verifier} & \textbf{Single} & \textbf{Two} & \textbf{Counting} & \textbf{Color} & \textbf{Position} & \textbf{Attribution} & \textbf{Overall} \\
    \midrule
    \rowcolor{Gray}
    \multicolumn{8}{l}{\text{VHS}} \\
    w/ $h_{1}$ Weighted XE Loss & 99.7 & 88.1 & 56.8 & 87.8 & 51.8 & 48.2  & 71.3 \\
    w/ $h_{5}$ Weighted XE Loss & \textbf{100.0} & 92.3 & 58.5 & 90.0 & 66.6 & 59.4 & 77.7 \\
    w/ $h_{9}$ Weighted XE Loss & \textbf{100.0} & 93.1 & 65.5 & 88.9 & 65.0 & 59.8 & 78.3 \\
    w/ $h_{19}$ Weighted XE Loss & 99.7 & 88.1 & 61.7 & 88.5 & 65.6 & 57.8 & 76.5  \\
    \cmidrule{1-8}
    w/ $h_{7}$ XE Loss & \textbf{100.0} & 90.9 & 59.5 & 88.1 & 61.6 & 60.4 & 76.3 \\
    w/ $h_{7}$ Focal Loss & \textbf{100.0} & 94.7 & 64.8 & 88.7 & \textbf{71.8} & 62.0 & 80.0  \\
    \rowcolor{OurColor}
    w/ $h_{7}$ Weighted XE Loss \textbf{(Ours)} & \textbf{100.0} & \textbf{95.7} & \textbf{66.5} & 88.9 & 69.8 & \textbf{63.8} & \textbf{80.5} \\
    w/ $h_{7}$ Weighted XE Loss + Qwen2-1.5B~\cite{team2024qwen2} & \textbf{100.0} & 94.1 & 62.0 & \textbf{90.4} & 65.0 & 61.0 & 78.4 \\
    \cmidrule{1-8}
    \rowcolor{Gray}
    \multicolumn{8}{l}{\text{MLLM w/ CLIP}} \\
    Generated Data w/ XE Loss & \textbf{100.0} & 94.1 & 64.0 & \textbf{90.2} & 63.0 & 62.0 & 78.5 \\
    Generated Data w/ Weighted XE Loss & \textbf{100.0} & 94.1 & 62.8 & 90.0 & 64.0 & \textbf{63.0} & 78.6 \\ 
    Original Data & \textbf{100.0} & 92.7 & \textbf{66.0} & 88.9 & 65.9 & 61.6 & \textbf{78.8} \\ 
    Original Data + Qwen2-1.5B~\cite{team2024qwen2} & 99.7 & \textbf{94.5} & 60.8 & 87.6 & \textbf{69.6} & 62.2 & \textbf{78.8} \\
    \bottomrule
  \end{tabular}
  }
  \label{tab:ablation}
  \vspace{-0.5cm}
\end{table*}

\tit{Overall Performance Analysis} Results are reported in Table~\ref{tab:geneval}. An analysis of the raw SANA-Sprint performance (first row) reveals substantial variation across the benchmark categories. While tasks such as counting, position, and attribute binding leave room for improvement, others, like single-object, two objects, and color, yield near-perfect accuracy, resulting in an overall score of 71.6\%. Across all time budgets, \ours consistently outperforms its CLIP-based counterpart, benefiting from being able to generate a larger pool of samples to select from, while maintaining comparable accuracy. In particular, \ours surpasses the baseline by 2.7\%, 1.7\%, and 0.5\% in the three time settings. 

Conversely, the MLLM w/ AE variant performs notably worse. AE latent features are perceptually richer thanks to the reconstruction pretraining objective of the autoencoder, but semantically weaker. As a result, these representations would likely require a more sophisticated architecture for semantic feature extraction. In practice, these features behave more like compressed, perceptual pixel-space representations rather than meaningful semantic embeddings.
In contrast, \ours leverages hidden-layer activations directly conditioned on the generation prompt, yielding much stronger semantic alignment than AE latents. This allows \ours to entirely remove the vision encoder while maintaining effective alignment with the LLM space through a lightweight MLP.

\begin{figure*}[t]
    \centering
    \includegraphics[width=0.94\linewidth]{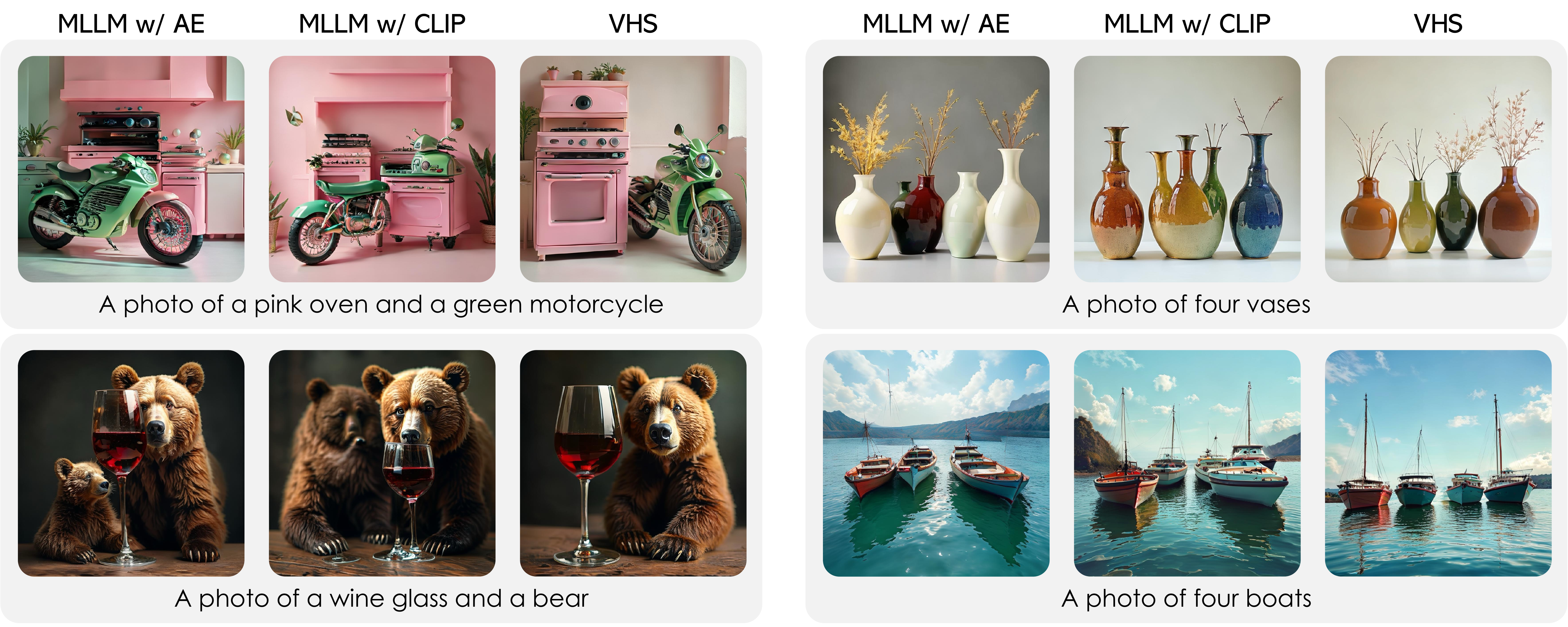}
    \vspace{-0.2cm}
    \caption{Visual comparison of the best pick images by different verifiers for GenEval-generated images.}
    \label{fig:qualitatives}
    \vspace{-0.4cm}
\end{figure*}

\tit{Category-wise Analysis} We observe the largest gains in categories that require generation over multiple objects. Specifically, \ours achieves up to a 3\% lead in the attribute binding (at 550 ms), 5.2\% in position (550\,ms), and up to 3\% in the two objects category (1100\,ms), indicating that \ours effectively distinguishes multiple objects and captures their spatial relationships and attributes. Conversely, the AE-equipped MLLM attains comparable, and in some cases superior, performance in the color category (89.6\%, 90.8\%, and 89.8\% across all the budgets). This can be attributed to the nature of AE latent features, where color information is more easily captured due to their perceptual rather than semantic representation space.
Moreover, the single-object category shows saturated values across all time windows and verification options, suggesting that on the simplest task proposed by the benchmark, the generator itself is good enough to yield almost perfect scores. To qualitatively validate \ours, Fig.~\ref{fig:qualitatives} presents samples from GenEval showcasing best picks from \ours, MLLM w/ CLIP and MLLM w/ AE, where \ours better identifies the best samples with equal inference time.
\tit{Multi-step baselines}
As additional evidence of the effectiveness of \ours, we compare against baselines that rely on multiple denoising steps. Consistent with prior findings~\cite{ma2025inference}, Best-of-N configurations outperform multi-step counterparts, both for the natively few-step SANA-Sprint and for its multi-step variant, SANA-1.5 (+9.5 and +1.5\% overall at 1650\,ms).

\subsection{Ablation Studies}
To assess the design space and justify our modeling choices, we perform a systematic ablation study on GenEval with SANA-Sprint, varying (i) the DiT layer from which visual latents are extracted, (ii) the LLM backbone, (iii) the loss function in the verifier fine-tuning stage, and (iv) the training data. The first two hyperparameters jointly determine both the final accuracy of \ours and the latency-accuracy trade-off of the verifier, directly impacting its adoption in real-world deployments. To enable a fair comparison, all evaluations are constrained to a fixed computational wall-time budget of 1100\,ms, with results reported in Table~\ref{tab:ablation}.

\tit{Ablation on Different DiT Layers} 
The behavior of \ours is tightly coupled to the visual information encoded in DiT layers, which capture varying levels of semantics and detail~\cite{kim2025seg4diff}. This induces a trade-off between expressivity and computational cost: deeper layers are more expensive to evaluate yet can provide richer semantic representations, while shallower layers are significantly cheaper but may encode weaker semantics. Crucially, this trade-off is not monotonic, as the final layers lie closest to the autoencoder reconstruction space, prioritizing perceptual fidelity over explicit semantic structures. As in Table~\ref{tab:geneval}, this proves suboptimal, highlighting the importance of selecting an appropriate latent depth.

We compare \ours trained with features extracted from layer $h_7$ against variants spanning a broad range of depths, from extreme layers $h_1$ and $h_{19}$ to intermediate ones $h_5$ and $h_9$ (approximately 25\% and 45\% of the depth of a 20-layer DiT, respectively). Extreme layers prove substantially detrimental: $h_1$ suffers from proximity to the noisy input regime, yielding unstable representations, while $h_{19}$ produces features dominated by perceptual reconstruction cues, consistent with the poor performance of the MLLM w/ AE baseline, confirming that both extremes provide weak semantic signals for verification. Among intermediate layers, $h_7$ yields consistent gains of 2.8\% and 2.2\% over $h_5$ and $h_9$, respectively, on the overall GenEval score. $h_5$ is substantially penalized on semantically demanding categories such as counting (-8\% compared to $h_7$) and attribution (-4.4\%), indicating that features extracted too early provide an insufficiently mature semantic signal. Conversely, the higher cost of $h_9$ limits sample exploration under fixed wall-time budget, ultimately hurting overall performance.

\tit{Ablation on Different Losses} Employing standard XE degrades performance compared to our weighted XE objective, with a drop of 4.2\% in the overall score for \ours and 0.1\% for the MLLM w/ CLIP baseline. This trend is consistent with the label imbalance in the SANA-Sprint training data, where positive examples account for approximately 63\% of the samples: an unweighted loss biases the verifier toward the majority (positive) class, impairing its ability to reject incorrect generations. Weighted XE counteracts this bias and yields systematic gains across most GenEval categories. Another solution to class imbalance is the focal loss~\cite{lin2017focal}, which down-weights well-classified examples and focuses the training signal on harder, misclassified samples, improving robustness on underrepresented and challenging cases. Indeed, training \ours with focal loss leads to a +3.7\% overall improvement over vanilla XE, confirming the importance of loss functions that explicitly account for class imbalance.

\tit{Ablation on Different LLM Backbones}
A similar trade-off arises on the language side: larger LLMs generally offer stronger reasoning capabilities and better alignment with task instructions, but at the cost of increased inference latency and memory footprint. Differently, increasing the LLM capacity by replacing Qwen2.5-0.5B with Qwen2-1.5B yields only marginal, and sometimes negative, gains under the same wall-time budget. This suggests that the primary bottleneck lies in the quality and depth of the visual representations rather than in the reasoning power of the language model, and that investing computation in better visual latents and appropriate losses is more beneficial than scaling up the~LLM.

\tit{Training Data}
We further analyze the MLLM w/ CLIP baseline to assess the impact of synthetic fine-tuning data. Notably, MLLM w/ CLIP shows no meaningful improvement, and even slight degradation ($-0.3\%$ and $-0.2\%$), when trained on generated rather than original data. This suggests that synthetic pairs provide little benefit for models not leveraging internal DiT latents. Therefore, the gains observed with \ours stem not from extra synthetic supervision but from its architectural design, which leverages DiT-layer latents and tailored loss functions, demonstrating the effectiveness of our verifier over generic MLLM-based baselines. Moreover, we refer the reader to the supplementary material for an analysis of the generated data quality.

\subsection{Generalization to Other Generators}
Finally, we provide an analysis on \ours when applied to a different single-step generator, in particular PixArt-$\alpha$-DMD~\cite{chenpixart,Yin_2024_CVPR}. Results are reported in Table~\ref{tab:pixart}.

\tit{Experimental Setting} Following the methodology from our SANA-Sprint analysis, we evaluate three verification approaches: the conventional pipeline using MLLM w/ CLIP features, direct verification on latent autoencoder features (MLLM w/ AE), and \ours operating on intermediate DiT activations from layer 13. Based on the previous ablation study, we train \ours with weighted XE loss and benchmark against the MLLM w/ CLIP variant trained on the original dataset, both identified as optimal configurations in our ablations.

\begin{table}[t]
  \centering
  \setlength{\tabcolsep}{.25em}
  \caption{Verification latency and GenEval~\cite{ghosh2023geneval} accuracy scores for Best-of-$N$ generation with PixArt-$\alpha$~\cite{chenpixart,Yin_2024_CVPR} under equivalent computational budgets defined by MLLM w/ CLIP.}
  \vspace{-0.15cm}
  \resizebox{\linewidth}{!}{
  \begin{tabular}{l c c c c c c c c c}
    \toprule
     & & \multicolumn{2}{c}{\textbf{Verification Time}} & & \multicolumn{5}{c}{\textbf{GenEval Overall (\%)}} \\
     \cmidrule(lr){3-4} \cmidrule(lr){6-10}
     & & \textbf{t (ms)} & \textbf{t savings (\%)} & & $\mathsf{Bo2}$ & $\mathsf{Bo3}$ & $\mathsf{Bo4}$ & $\mathsf{Bo5}$ & $\mathsf{Bo6}$ \\
    \midrule
    MLLM w/ CLIP & & 145 & - & & \textbf{43.7} & 44.7 & 45.1 & 45.6 & \textbf{46.9 }\\
    MLLM w/ AE & & 165 & $-$14.0 & & 41.0 & 41.0 & 41.9 & 42.3 & 41.6 \\
    \midrule
        \rowcolor{OurColor}
    \ours on $h_{13}$ & & \textbf{76} & \textbf{48.0} & & 43.0 & \textbf{45.2} & \textbf{45.5} & \textbf{46.1} & 46.4 \\
    \bottomrule
  \end{tabular}
  }
  \label{tab:pixart}
  \vspace{-0.4cm}
\end{table}

\tit{Latency Estimation} Inference-time analysis reveals that \ours achieves a $48\%$ speedup compared to MLLM w/ CLIP. In contrast, MLLM w/ AE offers no computational advantage over the CLIP baseline, as the generator autoencoder uses a low compression ratio that produces significantly more visual tokens for the LLM, negating any gains from bypassing latent decoding and CLIP encoding steps.

\tit{GenEval Performance} On GenEval, we evaluate performance under matched budgets, corresponding to sampling and scoring between two and six candidates with a CLIP-based verifier. Thanks to its lower latency, \ours attains the best results in the $\mathsf{Bo3}$ (45.2), $\mathsf{Bo4}$ (45.5), and $\mathsf{Bo5}$ (46.1) settings, while remaining comparable in the $\mathsf{Bo2}$ and $\mathsf{Bo6}$.

\vspace{-0.1cm}
\section{Conclusion}
\vspace{-0.1cm}
In this work, we introduced \ours, a verifier for inference-time scaling that directly aligns the latent representations of a DiT-based image generator with a large language model. By operating entirely in latent space, \ours eliminates part of the generation process, as well as the decode-re-encode overhead of standard MLLM-based verifiers, ultimately yielding better performance under the same inference-time budget.

\subsection*{Acknowledgments}
We acknowledge CINECA for the availability of high-performance computing resources under the ISCRA initiative, and for funding Evelyn Turri's PhD. This work has been supported by the EU Horizon projects ``ELIAS'' (GA No. 101120237) and ``ELLIOT'' (GA No. 101214398), and by the EuroHPC JU project  ``MINERVA'' (GA No. 101182737).

{
    \small
    \bibliographystyle{ieeenat_fullname}
    \bibliography{bibliography}

@string{cvpr     = {CVPR}}

@string{nips     = {NeurIPS}}

@string{nipsw    = {NeurIPS Workshops}}

@string{iccv     = {ICCV}}

@string{eccv     = {ECCV}}

@string{iclr     = {ICLR}}

@string{bmvc     = {BMVC}}

@string{icml     = {ICML}}

@string{emnlp    = {EMNLP}}

@string{miccai   = {MICCAI}}

@string{aclf     = {ACL Findings}}

@inproceedings{ho2020denoising,
  title={{Denoising Diffusion Probabilistic Models}},
  author={Ho, Jonathan and Jain, Ajay and Abbeel, Pieter},
  booktitle=nips,
  year={2020}
}

@inproceedings{lipman2022flow,
  title={{Flow Matching for Generative Modeling}},
  author={Lipman, Yaron and Chen, Ricky TQ and Ben-Hamu, Heli and Nickel, Maximilian and Le, Matt},
  booktitle=iclr,
  year={2023}
}

@InProceedings{Peebles_2023_ICCV,
    author    = {Peebles, William and Xie, Saining},
    title     = {{Scalable Diffusion Models with Transformers}},
    booktitle = iccv,
    year      = {2023},
}

@InProceedings{li2025reflect,
  title={{Reflect-DiT: Inference-Time Scaling for Text-to-Image Diffusion Transformers via In-Context Reflection}},
  author={Li, Shufan and Kallidromitis, Konstantinos and Gokul, Akash and Koneru, Arsh and Kato, Yusuke and Kozuka, Kazuki and Grover, Aditya},
  booktitle=iccv,
  year={2025}
}

@InProceedings{ghosh2023geneval,
  title={{GenEval: An Object-Focused Framework for Evaluating Text-to-Image Alignment
}},
  author={Ghosh, Dhruba and Hajishirzi, Hannaneh and Schmidt, Ludwig},
  booktitle=nips,
  year={2023}
}

@inproceedings{rombach2022high,
  title={{High-resolution image synthesis with latent diffusion models}},
  author={Rombach, Robin and Blattmann, Andreas and Lorenz, Dominik and Esser, Patrick and Ommer, Bj{\"o}rn},
  booktitle=cvpr,
  year={2022}
}

@inproceedings{caffagni2024revolution,
  title={{The Revolution of Multimodal Large Language Models: A Survey}},
  author={Caffagni, Davide and Cocchi, Federico and Barsellotti, Luca and Moratelli, Nicholas and Sarto, Sara and Baraldi, Lorenzo and Cornia, Marcella and Cucchiara, Rita},
  booktitle=aclf,
  year={2024}
}

@inproceedings{Yin_2024_CVPR,
    author    = {Yin, Tianwei and Gharbi, Micha\"el and Zhang, Richard and Shechtman, Eli and Durand, Fr\'edo and Freeman, William T. and Park, Taesung},
    title     = {{One-step Diffusion with Distribution Matching Distillation}},
    booktitle = cvpr,
    year      = {2024},
}

@inproceedings{chen2025sana,
  title={{SANA-Sprint: One-Step Diffusion with Continuous-Time Consistency Distillation}},
  author={Chen, Junsong and Xue, Shuchen and Zhao, Yuyang and Yu, Jincheng and Paul, Sayak and Chen, Junyu and Cai, Han and Han, Song and Xie, Enze},
  booktitle=iccv,
  year={2025}
}

@inproceedings{song2020denoising,
  title={Denoising Diffusion Implicit Models},
  author={Song, Jiaming and Meng, Chenlin and Ermon, Stefano},
  booktitle=iclr,
  year={2021}
}

@inproceedings{sauer2024adversarial,
  title={{Adversarial Diffusion Distillation}},
  author={Sauer, Axel and Lorenz, Dominik and Blattmann, Andreas and Rombach, Robin},
  booktitle=eccv,
  year={2024},
}

@inproceedings{lin2024evaluating,
        title={{Evaluating Text-to-Visual Generation with Image-to-Text Generation}},
        author={Lin, Zhiqiu and Pathak, Deepak and Li, Baiqi and Li, Jiayao and Xia, Xide and Neubig, Graham and Zhang, Pengchuan and Ramanan, Deva},
        booktitle=eccv,
        year={2024}
}

@article{xu2024visionreward,
  title={{VisionReward: Fine-Grained Multi-Dimensional Human Preference Learning for Image and Video Generation}},
  author={Xu, Jiazheng and Huang, Yu and Cheng, Jiale and Yang, Yuanming and Xu, Jiajun and Wang, Yuan and Duan, Wenbo and Yang, Shen and Jin, Qunlin and Li, Shurun and others},
  journal={arXiv preprint arXiv:2412.21059},
  year={2024}
}

@InProceedings{snell2024scaling,
  title={{Scaling LLM Test-Time Compute Optimally can be More Effective than Scaling Model Parameters}},
  author={Snell, Charlie and Lee, Jaehoon and Xu, Kelvin and Kumar, Aviral},
  booktitle=iclr,
  year={2025}
}

@inproceedings{wu2025inference,
  title={{Inference Scaling Laws: An Empirical Analysis of Compute-Optimal Inference for LLM Problem-Solving}},
  author={Wu, Yangzhen and Sun, Zhiqing and Li, Shanda and Welleck, Sean and Yang, Yiming},
  booktitle=iclr,
  year={2025}
}

@article{guan2025rstar,
  title={rStar-Math: Small LLMs Can Master Math Reasoning with Self-Evolved Deep Thinking},
  author={Guan, Xinyu and Zhang, Li Lyna and Liu, Yifei and Shang, Ning and Sun, Youran and Zhu, Yi and Yang, Fan and Yang, Mao},
  journal={arXiv preprint arXiv:2501.04519},
  year={2025}
}

@article{xie2025sana,
  title={{SANA 1.5: Efficient Scaling of Training-Time and Inference-Time Compute in Linear Diffusion Transformer}},
  author={Xie, Enze and Chen, Junsong and Zhao, Yuyang and Yu, Jincheng and Zhu, Ligeng and Lin, Yujun and Zhang, Zhekai and Li, Muyang and Chen, Junyu and Cai, Han and others},
  journal={arXiv preprint arXiv:2501.18427},
  year={2025}
}

@article{kim2025inference,
  title={{Inference-Time Scaling for Flow Models via Stochastic Generation and Rollover Budget Forcing}},
  author={Kim, Jaihoon and Yoon, Taehoon and Hwang, Jisung and Sung, Minhyuk},
  journal={arXiv preprint arXiv:2503.19385},
  year={2025}
}

@article{singh2025code,
  title={{CoDe: Blockwise Control for Denoising Diffusion Models}},
  author={Singh, Anuj and Mukherjee, Sayak and Beirami, Ahmad and Jamali-Rad, Hadi},
  journal={arXiv preprint arXiv:2502.00968},
  year={2025}
}

@InProceedings{ma2025inference,
  title={{Inference-Time Scaling for Diffusion Models beyond Scaling Denoising Steps}},
  author={Ma, Nanye and Tong, Shangyuan and Jia, Haolin and Hu, Hexiang and Su, Yu-Chuan and Zhang, Mingda and Yang, Xuan and Li, Yandong and Jaakkola, Tommi and Jia, Xuhui and others},
  booktitle=cvpr,
  year={2025}
}

@article{luo2023latent,
  title={Latent consistency models: Synthesizing high-resolution images with few-step inference},
  author={Luo, Simian and Tan, Yiqin and Huang, Longbo and Li, Jian and Zhao, Hang},
  journal={arXiv preprint arXiv:2310.04378},
  year={2023}
}

@inproceedings{radford2021learning,
  title={{Learning Transferable Visual Models From Natural Language Supervision}},
  author={Radford, Alec and Kim, Jong Wook and Hallacy, Chris and Ramesh, Aditya and Goh, Gabriel and Agarwal, Sandhini and Sastry, Girish and Askell, Amanda and Mishkin, Pamela and Clark, Jack and others},
  booktitle=icml,
  year={2021},
}

@inproceedings{liu2023visual,
  title={{Visual Instruction Tuning}},
  author={Liu, Haotian and Li, Chunyuan and Wu, Qingyang and Lee, Yong Jae},
  booktitle=nips,
  year={2023}
}

@inproceedings{laurenccon2024building,
  title={Building and better understanding vision-language models: insights and future directions},
  author={Lauren{\c{c}}on, Hugo and Marafioti, Andr{\'e}s and Sanh, Victor and Tronchon, L{\'e}o},
  booktitle=nipsw,
  year={2024}
}

@article{bai2025qwen2,
  title={{Qwen2.5-VL Technical Report}},
  author={Bai, Shuai and Chen, Keqin and Liu, Xuejing and Wang, Jialin and Ge, Wenbin and Song, Sibo and Dang, Kai and Wang, Peng and Wang, Shijie and Tang, Jun and others},
  journal={arXiv preprint arXiv:2502.13923},
  year={2025}
}

@inproceedings{tong2024cambrian,
  title={{Cambrian-1: A Fully Open, Vision-Centric Exploration of Multimodal LLMs}},
  author={Tong, Peter and Brown, Ellis and Wu, Penghao and Woo, Sanghyun and IYER, Adithya Jairam Vedagiri and Akula, Sai Charitha and Yang, Shusheng and Yang, Jihan and Middepogu, Manoj and Wang, Ziteng and others},
  booktitle=nips,
  year={2024}
}

@inproceedings{goodfellow2014generative,
  title={{Generative Adversarial Nets}},
  author={Goodfellow, Ian J and Pouget-Abadie, Jean and Mirza, Mehdi and Xu, Bing and Warde-Farley, David and Ozair, Sherjil and Courville, Aaron and Bengio, Yoshua},
  booktitle=nips,
  year={2014}
}

@inproceedings{ronneberger2015u,
  title={{U-Net: Convolutional Networks for Biomedical Image Segmentation}},
  author={Ronneberger, Olaf and Fischer, Philipp and Brox, Thomas},
  booktitle=miccai,
  year={2015},
}

@inproceedings{esser2024scaling,
  title={{Scaling Rectified Flow Transformers for High-Resolution Image Synthesis}},
  author={Esser, Patrick and Kulal, Sumith and Blattmann, Andreas and Entezari, Rahim and M{\"u}ller, Jonas and Saini, Harry and Levi, Yam and Lorenz, Dominik and Sauer, Axel and Boesel, Frederic and others},
  booktitle=icml,
  year={2024}
}

@article{team2025gemma,
  title={{Gemma 3 Technical Report}},
  author={Team, Gemma and Kamath, Aishwarya and Ferret, Johan and Pathak, Shreya and Vieillard, Nino and Merhej, Ramona and Perrin, Sarah and Matejovicova, Tatiana and Ram{\'e}, Alexandre and Rivi{\`e}re, Morgane and others},
  journal={arXiv preprint arXiv:2503.19786},
  year={2025}
}

@inproceedings{zhuo2025reflection,
  title={{From Reflection to Perfection: Scaling Inference-Time Optimization for Text-to-Image Diffusion Models via Reflection Tuning}},
  author={Zhuo, Le and Zhao, Liangbing and Paul, Sayak and Liao, Yue and Zhang, Renrui and Xin, Yi and Gao, Peng and Elhoseiny, Mohamed and Li, Hongsheng},
  booktitle=iccv,
  year={2025}
}

@inproceedings{chen2024deep,
  title={{Deep Compression Autoencoder for Efficient High-Resolution Diffusion Models}},
  author={Chen, Junyu and Cai, Han and Chen, Junsong and Xie, Enze and Yang, Shang and Tang, Haotian and Li, Muyang and Lu, Yao and Han, Song},
  booktitle=iclr,
  year={2025}
}

@inproceedings{he-etal-2024-videoscore,
    title = "{V}ideo{S}core: Building Automatic Metrics to Simulate Fine-grained Human Feedback for Video Generation",
    author = "He, Xuan  and
      Jiang, Dongfu  and
      Zhang, Ge  and
      Ku, Max  and
      Soni, Achint  and
      Siu, Sherman  and
      Chen, Haonan  and
      Chandra, Abhranil  and
      Jiang, Ziyan  and
      Arulraj, Aaran  and
      Wang, Kai  and
      Do, Quy Duc  and
      Ni, Yuansheng  and
      Lyu, Bohan  and
      Narsupalli, Yaswanth  and
      Fan, Rongqi  and
      Lyu, Zhiheng  and
      Lin, Bill Yuchen  and
      Chen, Wenhu",
    booktitle = emnlp,
    year = "2024",
}

@article{team2024qwen2,
  title={{Qwen2 Technical Report}},
  author={Team, Qwen and others},
  journal={arXiv preprint arXiv:2407.10671},
  year={2024}
}

@inproceedings{chenpixart,
  title={{PixArt-$\alpha$: Fast Training of Diffusion Transformer for Photorealistic Text-to-Image Synthesis}},
  author={Chen, Junsong and Jincheng, YU and Chongjian, GE and Yao, Lewei and Xie, Enze and Wang, Zhongdao and Kwok, James and Luo, Ping and Lu, Huchuan and Li, Zhenguo},
  booktitle=iclr,
  year=2024
}

@inproceedings{lin2017focal,
  title={{Focal Loss for Dense Object Detection}},
  author={Lin, Tsung-Yi and Goyal, Priya and Girshick, Ross and He, Kaiming and Doll{\'a}r, Piotr},
  booktitle=iccv,
  year={2017}
}

@inproceedings{liu2024improved,
  title={Improved baselines with visual instruction tuning},
  author={Liu, Haotian and Li, Chunyuan and Li, Yuheng and Lee, Yong Jae},
  booktitle=cvpr,
  year={2024}
}

@inproceedings{lin2024vila,
  title={{VILA: On Pre-training for Visual Language Models}},
  author={Lin, Ji and Yin, Hongxu and Ping, Wei and Molchanov, Pavlo and Shoeybi, Mohammad and Han, Song},
  booktitle=cvpr,
  year={2024}
}

@inproceedings{cherti2023reproducible,
  title={{Reproducible Scaling Laws for Contrastive Language-Image Learning}},
  author={Cherti, Mehdi and Beaumont, Romain and Wightman, Ross and Wortsman, Mitchell and Ilharco, Gabriel and Gordon, Cade and Schuhmann, Christoph and Schmidt, Ludwig and Jitsev, Jenia},
  booktitle=cvpr,
  year={2023}
}

@inproceedings{zhai2023sigmoid,
  title={{Sigmoid Loss for Language Image Pre-Training}},
  author={Zhai, Xiaohua and Mustafa, Basil and Kolesnikov, Alexander and Beyer, Lucas},
  booktitle=iccv,
  year={2023}
}

@InProceedings{Xu_2023_CVPR,
    author    = {Xu, Jiarui and Liu, Sifei and Vahdat, Arash and Byeon, Wonmin and Wang, Xiaolong and De Mello, Shalini},
    title     = {{Open-Vocabulary Panoptic Segmentation With Text-to-Image Diffusion Models}},
    booktitle = cvpr,
    year      = {2023},
}

@inproceedings{NEURIPS2023_0503f5dc,
 author = {Tang, Luming and Jia, Menglin and Wang, Qianqian and Phoo, Cheng Perng and Hariharan, Bharath},
 booktitle = nips,
 editor = {A. Oh and T. Naumann and A. Globerson and K. Saenko and M. Hardt and S. Levine},
 title = {{Emergent Correspondence from Image Diffusion}},
 year = {2023}
}

@InProceedings{PNVR_2023_ICCV,
    author    = {PNVR, Koutilya and Singh, Bharat and Ghosh, Pallabi and Siddiquie, Behjat and Jacobs, David},
    title     = {{LD-ZNet: A Latent Diffusion Approach for Text-Based Image Segmentation}},
    booktitle = iccv,
    month     = {October},
    year      = {2023},
}

@article{hendrycks2016gaussian,
  title={{Gaussian Error Linear Units (GELUs)}},
  author={Hendrycks, D},
  journal={arXiv preprint arXiv:1606.08415},
  year={2016}
}

@inproceedings{loshchilovdecoupled,
  title={{Decoupled Weight Decay Regularization}},
  author={Loshchilov, Ilya and Hutter, Frank},
  booktitle=iclr,
  year={2019}
}

@inproceedings{baraldi2025verifier,
  title={Verifier Matters: Enhancing Inference-Time Scaling for Video Diffusion Models},
  author={Baraldi, Lorenzo and Bucciarelli, Davide and Zeng, Zifan and Zhang, Chongzhe and Zhang, Qunli and Cornia, Marcella and others},
  booktitle=bmvc,
  year={2025}
}

@inproceedings{kim2025seg4diff,
  title={{Seg4Diff: Unveiling Open-Vocabulary Segmentation in Text-to-Image Diffusion Transformers}},
  author={Kim, Chaehyun and Shin, Heeseong and Hong, Eunbeen and Yoon, Heeji and Arnab, Anurag and Seo, Paul Hongsuck and Hong, Sunghwan and Kim, Seungryong},
  booktitle=nips,
  year={2025}
}

@article{wang2026moonworks,
  title={{Moonworks Lunara Aesthetic Dataset}},
  author={{Wang, Yan and Abdullah, MM and Hassan, Partho and Hassan, Sabit}},
  journal={arXiv preprint arXiv:2601.07941},
  year={2026}
}

@inproceedings{hessel2021clipscore,
  title={{CLIPScore: A Reference-free Evaluation Metric for Image Captioning}},
  author={Hessel, Jack and Holtzman, Ari and Forbes, Maxwell and Le Bras, Ronan and Choi, Yejin},
  booktitle=emnlp,
  year={2021}
}

@inproceedings{hulora,
  title={{LoRA: Low-Rank Adaptation of Large Language Models}},
  author={Hu, Edward J and Wallis, Phillip and Allen-Zhu, Zeyuan and Li, Yuanzhi and Wang, Shean and Wang, Lu and Chen, Weizhu and others},
  booktitle=iclr,
  year={2022}
}

@misc{blackforestlabs2024fluxschnell,
  author       = {{Black Forest Labs}},
  title        = {{FLUX.1 Schnell}},
  year         = {2024},
  howpublished = {\url{https://blackforestlabs.ai}},
  note         = { }
}

@misc{stabilityai2024sd3_5_large_turbo,
  author       = {{Stability AI}},
  title        = {{Stable Diffusion 3.5 Large Turbo}},
  year         = {2024},
  howpublished = {\url{https://huggingface.co/stabilityai/stable-diffusion-3.5-large-turbo}},
  note         = { },
  organization = {Stability AI}
}
}

\clearpage
\maketitlesupplementary

\appendix

\section{Additional Implementation Details}
\subsection{Dataset Construction}
We build datasets pairing DiT embeddings with task-specific targets: captions for the alignment phase and binary labels for fine-tuning. In the following, we detail this generation process for both training phases of \ours.

\tit{Alignment Dataset} Due to the noisiness of the image-text couples of the LLaVA pre-training dataset~\cite{liu2023visual}, we use Gemma3-4B~\cite{team2025gemma} to produce refined captions as prompts. We then generate images, extract the corresponding DiT embeddings, and re-caption the outputs to guarantee alignment.

\tit{Verifier Fine-Tuning Dataset} We leverage the prompt set from ReflectDiT~\cite{li2025reflect} to synthesize a new image corpus, generating 20 variants per prompt and extracting their internal activations. To establish ground-truth labels for these samples, we implement an automated annotation pipeline using Gemma3-4B. Specifically, the LLM generates the metadata required for the GenEval~\cite{ghosh2023geneval} verification pipeline by first classifying the prompt tag (\eg, single object, two objects, counting, position, colors, attribute binding) and subsequently deriving the specific inclusion and exclusion lists. Finally, the GenEval verifier processes the synthesized images with this metadata to assign a binary label to each sample. The LLM prompts for data generation are shown~in~Fig.~\ref{fig:caption_prompt}.

\subsection{Architectural Details}
\label{supp:details}
\tinytit{Hidden Features Extraction} 
We extract hidden layer features from the output of the $\ell^\star$ layer and normalize them using mean and variance statistics pre-computed over a 50k-example subset of the alignment training set. For SANA-Sprint at standard resolution, this process yields a representation of 1024 spatial features, each with a hidden dimensionality of 2240. Similarly, PixArt-$\alpha$-DMD produces a feature map of the same spatial size (1024 features), but with a smaller hidden dimensionality of 1152.  

\tit{AE Features Extraction} AE features are extracted by flattening the output latents of the generator across spatial dimensions. For SANA-Sprint, the deep compression autoencoder reduces spatial resolution by 32$\times$. For a standard 1024$\times$1024 input resolution, this produces a 32$\times$32 feature grid with a dimensionality of 32, which, after flattening and projection, results in 1024 image tokens as input to the LLM. Conversely, for PixArt-$\alpha$-DMD, the generator uses a standard autoencoder with KL loss~\cite{rombach2022high}. Given a 512$\times$512 input resolution, this process yields a 64$\times$64 feature grid, which we flatten and project into the LLM embedding space, producing 4096 image tokens. This high token count increases the inference latency of the LLM further rendering AE features unfeasible for our task.

\tit{Multimodal Connector} The connector module is implemented as a two-layer MLP with a GELU activation function~\cite{hendrycks2016gaussian}, with the first projection layer bringing the features to LLM-compatible dimensionality.

\tit{LLM Setup} The full prompt given to the verifier for evaluating images can be seen in Fig.~\ref{fig:caption_prompt}, allowing the evaluation procedure to be reproduced.

\subsection{Training Details}
\tinytit{Alignment Stage} The model is trained for one epoch on the 558k-sample alignment dataset. We use a total batch size of 512 distributed across 16 NVIDIA A100 GPUs. The learning rate follows a cosine schedule, peaking at $1 \times 10^{-3}$, following a warm-up phase lasting 3\% of the total training steps, with the AdamW~\cite{loshchilovdecoupled} optimizer. 

\tit{Verifier Fine-Tuning} On the other hand, the fine-tuning stage of our pipeline is carried out using a total batch-size of 64 on 8 NVIDIA A100 GPUs, a cosine learning rate scheduling with a maximum value of $2\times10^{-5}$, and the AdamW optimizer. We select the model yielding the best evaluation loss during a 10 epochs training. To address class imbalance in the SANA-Sprint training set, we employ a weighted cross-entropy loss, assigning weights of 0.37 to the positive class and 0.63 to the negative class, following sample distribution. Conversely, for the PixArt-$\alpha$-DMD experiments, equal weights are utilized to reflect the balanced class distribution of the dataset. Moreover, the focal loss ablation study follows the formulation proposed in~\cite{lin2017focal}:
\begin{equation}
\label{eq:focal}
\text{FL}(p_{\text{t}}) = -\alpha_{\text{t}}(1 - p_{\text{t}})^{\gamma} \log(p_{\text{t}}),
\end{equation}
where $p_{\text{t}}$ represents the estimated probability of the model for the ground-truth class, $\alpha_{\text{t}}$ is a weighting factor assigned to each class to counteract class imbalance, and $\gamma$ modulates the loss function by exponentially down-weighting ``easy'' examples, forcing the model to concentrate on ``hard'', misclassified examples that contribute the most to the training error. We set $\alpha$ as the class imbalance factor and $\gamma$ to $2$.

\begin{table*}[t]
  \centering
  \setlength{\tabcolsep}{1.em}
  \caption{Accuracy (\%) of SANA-Sprint~\cite{chen2025sana} on the GenEval benchmark~\cite{ghosh2023geneval}. Ablation on score methodology on a time budget of 1100\,ms.}
  \vspace{-0.3cm}
  \resizebox{0.98\linewidth}{!}{
  \begin{tabular}{l c c c c c c c}
    \toprule
     \textbf{Verifier} & \textbf{Single} & \textbf{Two} & \textbf{Counting} & \textbf{Color} & \textbf{Position} & \textbf{Attribution} & \textbf{Overall} \\
    \midrule
    \rowcolor{Gray}
    \multicolumn{8}{l}{\text{VHS}} \\
    w/ $h_{7}$ No Scoring & 99.3 & 90.7 & 57.5 & 87.5 & 63.2 & 52.6 & 74.7 \\
    \rowcolor{OurColor}
    w/ $h_{7}$ Token Probability Scoring \textbf{(Ours)} & \textbf{100.0} & \textbf{95.7} & \textbf{66.5} & 88.9 & \textbf{69.8} & \textbf{63.8} & \textbf{80.5} \\
    \cmidrule{1-8}
    \rowcolor{Gray}
    \multicolumn{8}{l}{\text{MLLM w/ CLIP}} \\
    w/ No Scoring & 99.5 & 91.1 & 58.3 & 88.3 & 62.0 & 54.8 & 75.3 \\
    w/ Token Probability Scoring & \textbf{100.0} & 92.7 & 66.0 & 88.9 & 65.9 & 61.6 & 78.8 \\ 
    \cmidrule{1-8}
    \rowcolor{Gray}
    \multicolumn{8}{l}{\text{MLLM w/ AE}} \\
    w/ No Scoring & 99.7 & 87.5 & 56.5 & 88.7 & 55.0 & 49.2 & 72.2 \\
    w/ Token Probability Scoring & 99.7 & 90.7 & 61.3 & \textbf{90.8} & 59.6 & 49.3 & 74.7 \\ 
    \bottomrule
  \end{tabular}
  }
  \label{tab:sup_ablation}
  \vspace{-0.15cm}
\end{table*}
\begin{table*}[t]
  \centering
  \setlength{\tabcolsep}{1em}
  \caption{GenEval benchmark~\cite{ghosh2023geneval} results with compute-equivalent baselines.}
  \vspace{-0.3cm}
  \resizebox{\linewidth}{!}{
  \begin{tabular}{c c c c c c c c c c c c}
    \toprule
     \textbf{Budget} & \textbf{Generator} & \textbf{Steps} & \textbf{Verifier} & $\mathsf{BoN}$ &  \textbf{Single} & \textbf{Two} & \textbf{Count} & \textbf{Color} & \textbf{Pos} & \textbf{Attr} & \textbf{All} \\
    \midrule
    \multirow{3}{*}{550ms}
    & SD3.5-Turbo & 1 & MLLM w/ CLIP & $\mathsf{Bo1}$ & 83.0 & 29.9 & 35.5 & 61.1 & 11.4 & 13.6  & 37.4 \\
    & Flux-Schnell & 1 & MLLM w/ CLIP & $\mathsf{Bo1}$ & 99.0 & 90.3 & 60.5 & 82.8 & 28.0 & \textbf{58.4} & 68.9 \\
    \cmidrule{2-12}
     \rowcolor{OurColor}
     & Sana-Sprint & 1 & \textbf{\ours (Ours)} & $\mathsf{Bo4}$ & \textbf{100.0} & \textbf{93.9} & \textbf{61.5} & \textbf{90.6} & \textbf{66.2} & \textbf{58.4} & \textbf{78.1}\\
     \midrule
    \multirow{4}{*}{1100ms}
    & SD3.5-Turbo & 1 & MLLM w/ CLIP & $\mathsf{Bo2}$ & 90.1 & 35.6 & 37.8 & 63.8 & 13.8 & 16.6 & 41.2 \\
    & Flux-Schnell & 1 & MLLM w/ CLIP & $\mathsf{Bo2}$ & 99.8 & 94.3 & 65.8 & 84.7 & 34.4 & \textbf{65.0} & 73.2 \\
    & Flux-Schnell & 2 & MLLM w/ CLIP & $\mathsf{Bo2}$ & 99.0 & 91.7 & 61.3 & 80.9 & 28.4 & 57.4 & 68.9  \\
    \cmidrule{2-12}
     \rowcolor{OurColor}
     & SANA-Sprint & 1 & \textbf{\ours (Ours)} & $\mathsf{Bo9}$ & \textbf{100.0} & \textbf{95.7} & \textbf{66.5} & \textbf{88.9} & \textbf{69.8} & 63.8 & \textbf{80.5}\\
     \midrule
    \multirow{4}{*}{1650ms}
    & SD3.5-Turbo & 1 & MLLM w/ CLIP & $\mathsf{Bo3}$ & 91.3 & 42.6 & 40.3 & 64.8 & 16.6 & 19.4 & 44.1  \\
    & Flux-Schnell & 1 & MLLM w/ CLIP & $\mathsf{Bo3}$ & \textbf{100.0} & 95.2 & 65.8 & 84.7 & 39.8 & \textbf{66.0} & 74.7 \\
    & Flux-Schnell & 2 & MLLM w/ CLIP & $\mathsf{Bo3}$ & \textbf{100.0} & 94.3 & \textbf{68.0} & 84.3 & 36.0 & 62.4 & 73.3\\
    \cmidrule{2-12}
     \rowcolor{OurColor}
     & SANA-Sprint & 1 & \textbf{\ours (Ours)} & $\mathsf{Bo15}$ & \textbf{100.0} & \textbf{96.0} & 67.3 & \textbf{89.1} & \textbf{70.4} & 64.6 & \textbf{80.9} \\
    \bottomrule
  \end{tabular}
  }
  \label{tab:additional_baselines}
  \vspace{-0.35cm}
\end{table*}
\begin{figure}[t]
    \centering
    \includegraphics[width=0.99\linewidth]{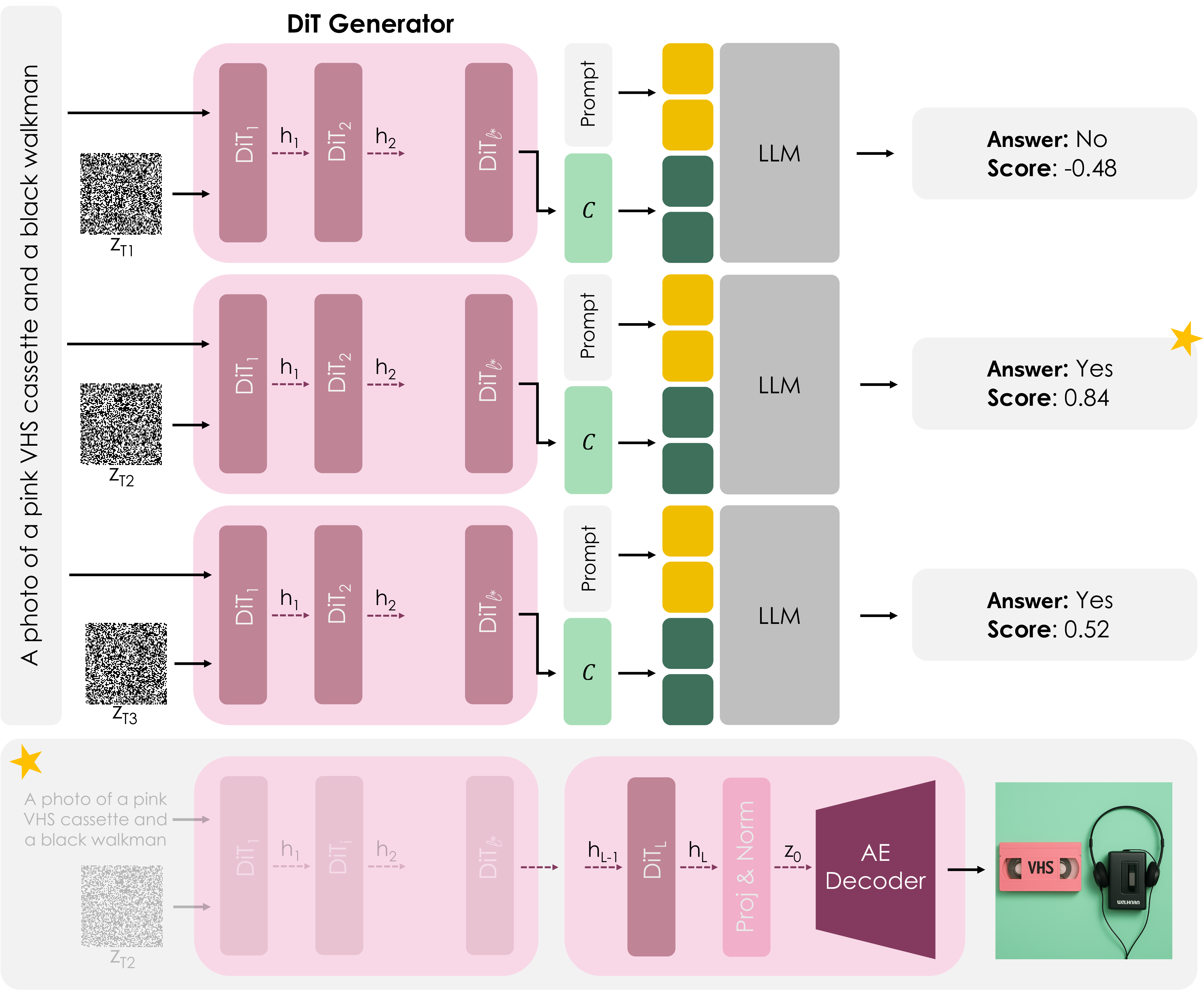}
    \vspace{-0.2cm}
    \caption{Efficient Best-of-N pipeline with \ours.}
    \label{fig:best-of-N}
    \vspace{-0.4cm}
\end{figure}
\begin{table*}[t]
  \centering
  \setlength{\tabcolsep}{1.4em}
  \caption{Accuracy (\%) of SANA-Sprint~\cite{chen2025sana} on the GenEval benchmark~\cite{ghosh2023geneval}, varying levels of  LoRA Fine-Tuning on a time budget of 1100\,ms for MLLM w/ CLIP and \ours.}
  \vspace{-0.3cm}
  \resizebox{\linewidth}{!}{
  \begin{tabular}{l c c c c c c c c}
    \toprule
     \textbf{Verifier} & \textbf{Single} & \textbf{Two} & \textbf{Counting} & \textbf{Color} & \textbf{Position} & \textbf{Attribution} & \textbf{Overall} & $\Delta$ \\
    \midrule
    \rowcolor{Gray}
    \multicolumn{9}{l}{\text{No LoRA Fine-Tuning}}\\
    MLLM w/ CLIP & \textbf{100.0} & 94.7 & 59.3 & \textbf{91.9} & 69.4 & 60.2 & 79.1 & \multirow{2}{*}{$+$1.4}\\
    \rowcolor{OurColor}
    \textbf{\ours} & \textbf{100.0} & \textbf{95.7} & \textbf{66.5} & 88.9 & \textbf{69.8} & \textbf{63.8} & \textbf{80.5} \\ 
    \cmidrule{1-9}
    \rowcolor{Gray}
    \multicolumn{9}{l}{\text{50\% LoRA Fine-Tuning}}\\
    MLLM w/ CLIP & \textbf{100.0} & \textbf{92.9} & 60.8 & \textbf{90.6} & 63.4 & 58.2 & 77.3  & \multirow{2}{*}{$+$1.8}\\
    \rowcolor{OurColor}
    \textbf{\ours} & \textbf{100.0} & 92.1 & \textbf{61.5} & 89.8 & \textbf{70.4} & \textbf{62.0} & \textbf{79.1} \\ 
    \cmidrule{1-9}
    \rowcolor{Gray}
    \multicolumn{9}{l}{\text{100\% LoRA Fine-Tuning}}\\
    MLLM w/ CLIP & 99.5 & 88.3 & 57.3 & 87.9 & 57.2 & \textbf{56.2}  & 73.9  & \multirow{2}{*}{$+$2.4}\\
    \rowcolor{OurColor}
    \textbf{\ours} & \textbf{100.0} & \textbf{91.1} & \textbf{59.5} & \textbf{89.4} & \textbf{64.4} & 56.0 & \textbf{76.3} \\ 
    \bottomrule
  \end{tabular}
  }
  \label{tab:lora}
  \vspace{-0.3cm}
\end{table*}
\begin{figure*}[t]
    \centering
    
    \begin{minipage}[t]{0.32\linewidth}
        \centering
        \includegraphics[width=\linewidth]{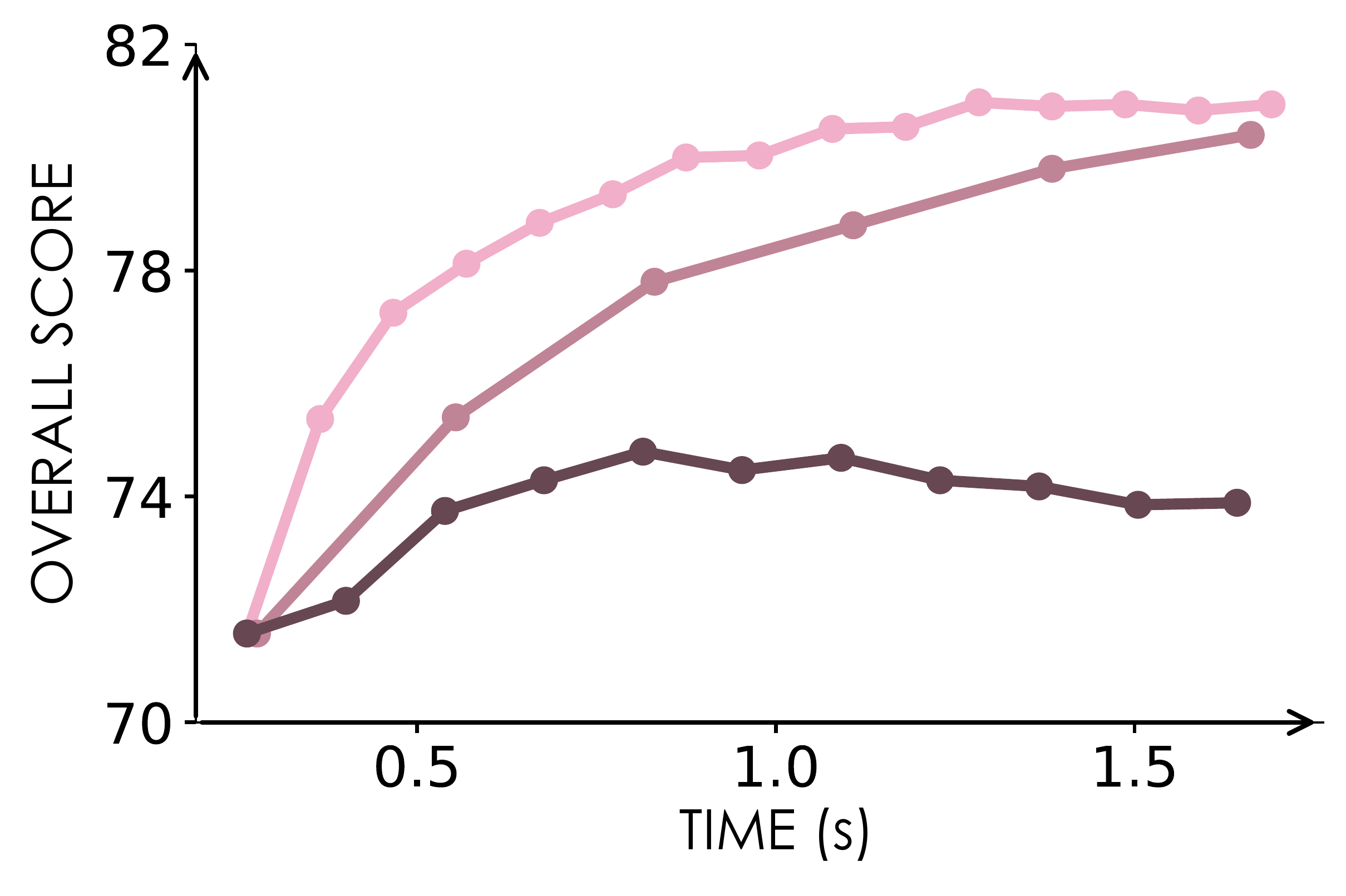}
        \vspace{-0.4cm}
    \end{minipage}
    \hfill
    \begin{minipage}[t]{0.32\linewidth}
        \centering
        \includegraphics[width=\linewidth]{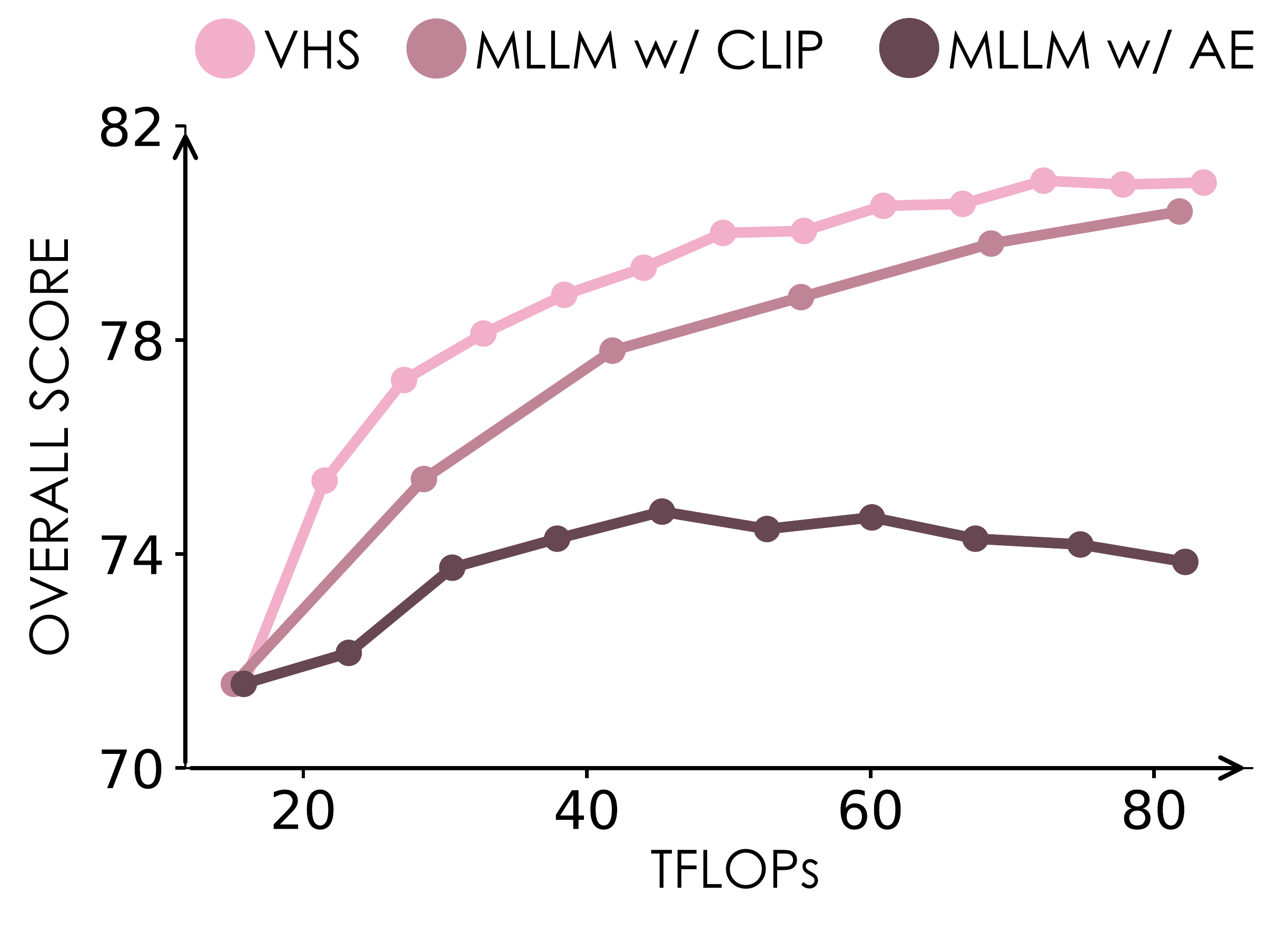}
        \vspace{-0.4cm}
    \end{minipage}
    \hfill
    \begin{minipage}[t]{0.32\linewidth}
        \centering
        \includegraphics[width=\linewidth]{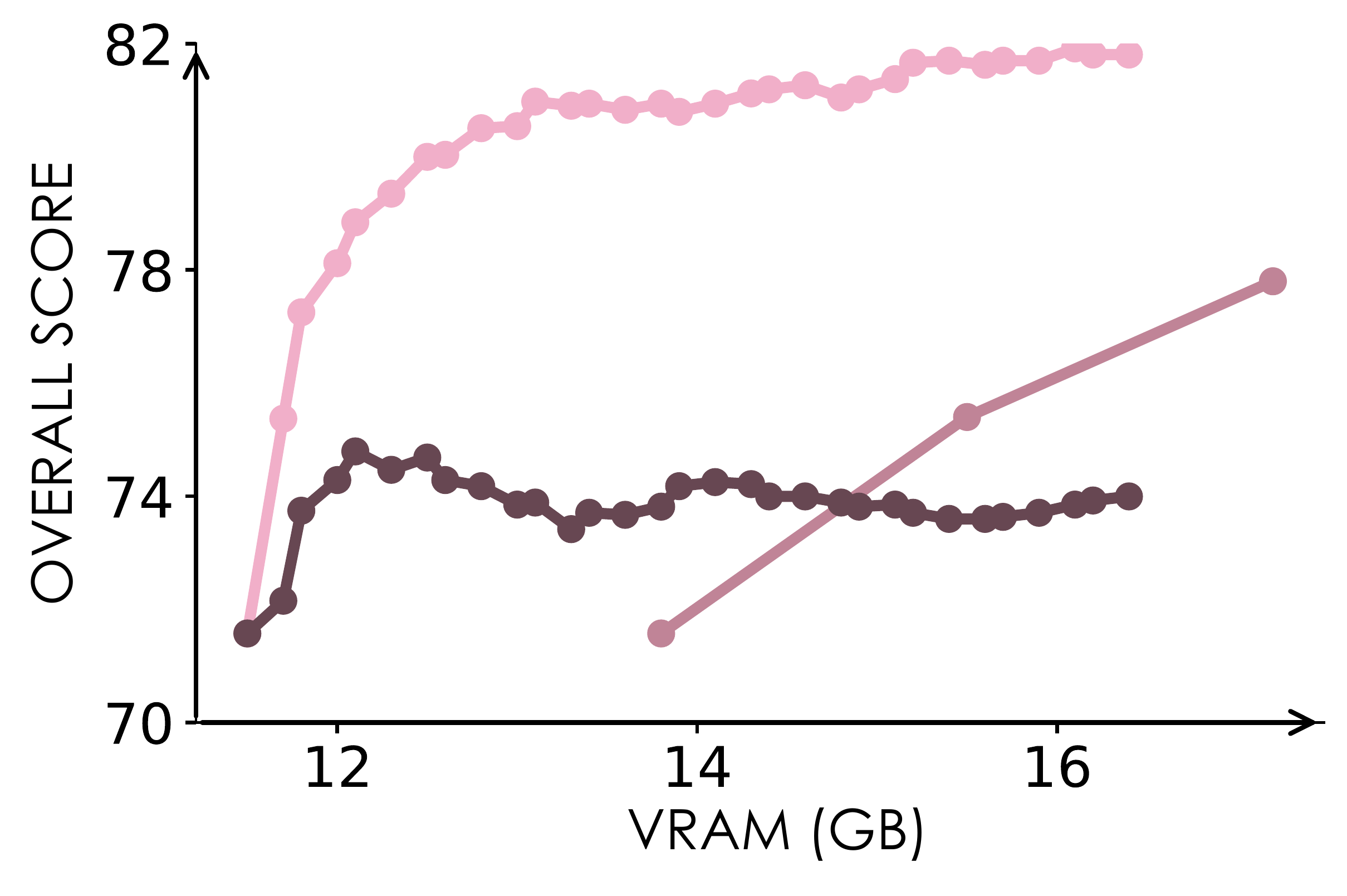}
        \vspace{-0.4cm}
    \end{minipage}
    
    \vspace{-0.4cm}
    \caption{Overall accuracy (\%) of SANA-Sprint~\cite{chen2025sana} on GenEval~\cite{ghosh2023geneval} across time (seconds) TFLOPs, and VRAM usage (GB).}
    \vspace{-0.4cm}
    \label{fig:three_plots}
\end{figure*}

\subsection{Inference Setup}
\label{supp:inference}

\tinytit{Score Computation} Following \cite{xie2025sana}, we convert binary verifier outputs into a continuous score. The score is defined as the predicted probability of the verifier for the sampled token, negated if it belongs to the negative class (\ie, ``no''). Otherwise (\ie, ``yes''), the probability is used as is.

\tit{Best-Of-N Setup} Fig.~\ref{fig:best-of-N} shows the Best-of-N setup: each candidate is passed through the first DiT layers and scored by the LLM, after which the highest-scoring candidate completes the remaining layers and is decoded to pixel space.

\tit{Inference and Evaluation Parameters} To run inference on the SANA-Sprint model, we stick to the standard resolution of 1024$\times$1024. We employ a CFG of 1.0, and run the generator in a single step. We average our results over 5 different seeds to get more stable estimations. Moreover, for PixArt-$\alpha$-DMD, the resolution is set 512$\times$512.

\section{Additional Analyses and Experiments}
\subsection{Data Quality Analysis}
To verify data quality for the alignment stage of \ours training, we evaluate the multimodal consistency of our generated image–caption pairs using CLIP-Score~\cite{hessel2021clipscore}. Our pipeline, which re-captions the original images, synthesizing new images, and re-captioning the outputs, produces pairs with a CLIP-Score of 73.5. Compared to the original dataset's score of 70.5, this demonstrates that our approach not only preserves semantic alignment, but actually improves image–text consistency over the original LLaVA alignment data.

\subsection{Additional Studies on the Continuous Score}
\begin{figure}[t]
    \centering
    \includegraphics[width=0.99\linewidth]{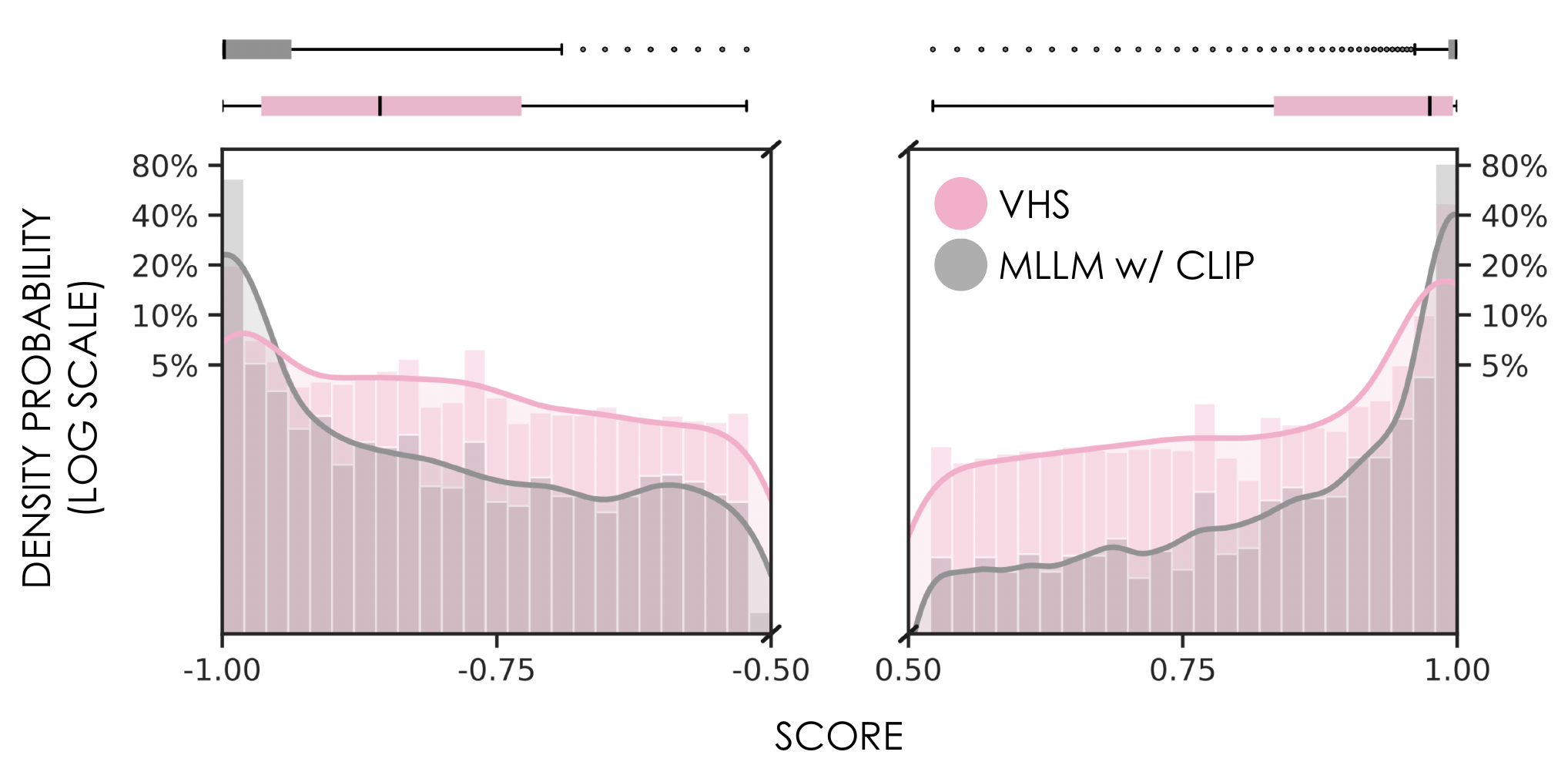}
    \vspace{-0.4cm}
    \caption{Score distribution comparison between MLLM w/ CLIP and \ours. Histograms display the frequency of scores within 5\% bins, with smoothed density curves overlaid in gray (MLLM w/ CLIP) and pink (\ours). Box plots above show the quartile ranges and distributional characteristics of each verifier.}
    \label{fig:supplementary_scores}
    \vspace{-0.4cm}
\end{figure}

\tinytit{Ablation on the Continuous Score} In Table~\ref{tab:sup_ablation}, we present an ablation study to validate the use of token probabilities as continuous scores for image selection. Specifically, we compare GenEval accuracies under an 1100\,ms budget with SANA-Sprint using two distinct strategies: (i) random selection among images classified as positive (``yes'') by the verifier, and (ii) selection of the highest-scoring image based on token probability, as described in Sec.~\ref{supp:inference}. We observe that utilizing probability scores consistently enhances accuracy across all categories and verifier types, yielding an overall improvement of up to 5.8\% for \ours. This suggests that token probabilities provide a granular measure of picture quality that binary decisions alone fail to capture.

\tinytit{Score Distribution Analysis}
In Fig.~\ref{fig:supplementary_scores}, we analyze the score distributions produced by \ours and MLLM w/ CLIP on the prompts of the GenEval benchmark, using 32 generations per prompt with SANA-Sprint.~The distributions reveal that \ours produces less extreme scores compared to MLLM w/ CLIP. Specifically, MLLM w/ CLIP assigns roughly 80\% of samples to the extreme score ranges ([0.95, 1.0] or [-1.0, -0.95]), indicating a strong tendency toward binary judgments. In contrast, \ours assigns only about 40\% and 20\% of samples to the positive and negative extremes, while distributing a larger proportion of scores toward the center of the range.~This more balanced distribution reflects the ability of \ours to assign more spread-out scores,~enabling finer-grained discrimination between samples of varying quality.

\subsection{Comparison with additional baselines}
In order to assess the trade-offs and resource allocation strategies within the Best-of-N pipeline under tight time budgets, in Table~\ref{tab:additional_baselines} we present additional comparisons between \ours and various baselines equipped with different generators, although not directly comparable. Specifically, we compare against FLUX.1-Schnell~\cite{blackforestlabs2024fluxschnell} and Stable Diffusion 3.5-Turbo\cite{stabilityai2024sd3_5_large_turbo}, exploring different Best-of-N and multi-step configurations under equivalent budgets of 650\,ms, 1100\,ms, and 1650\,ms. Overall, \ours achieves the best results across all evaluated time budgets, as configurations relying on larger samplers are forced to operate with too few denoising steps or single-shot selection, limiting their effectiveness. Finally, Fig.~\ref{fig:three_plots} presents an extended comparison of \ours and the two baselines, plotting overall GenEval score as a function of three computational budget metrics: inference time, TFLOPs, and VRAM usage (GB), demonstrating clear advantages for \ours across all three.
\begin{figure*}[t]
    \centering
    \includegraphics[width=0.98\linewidth]{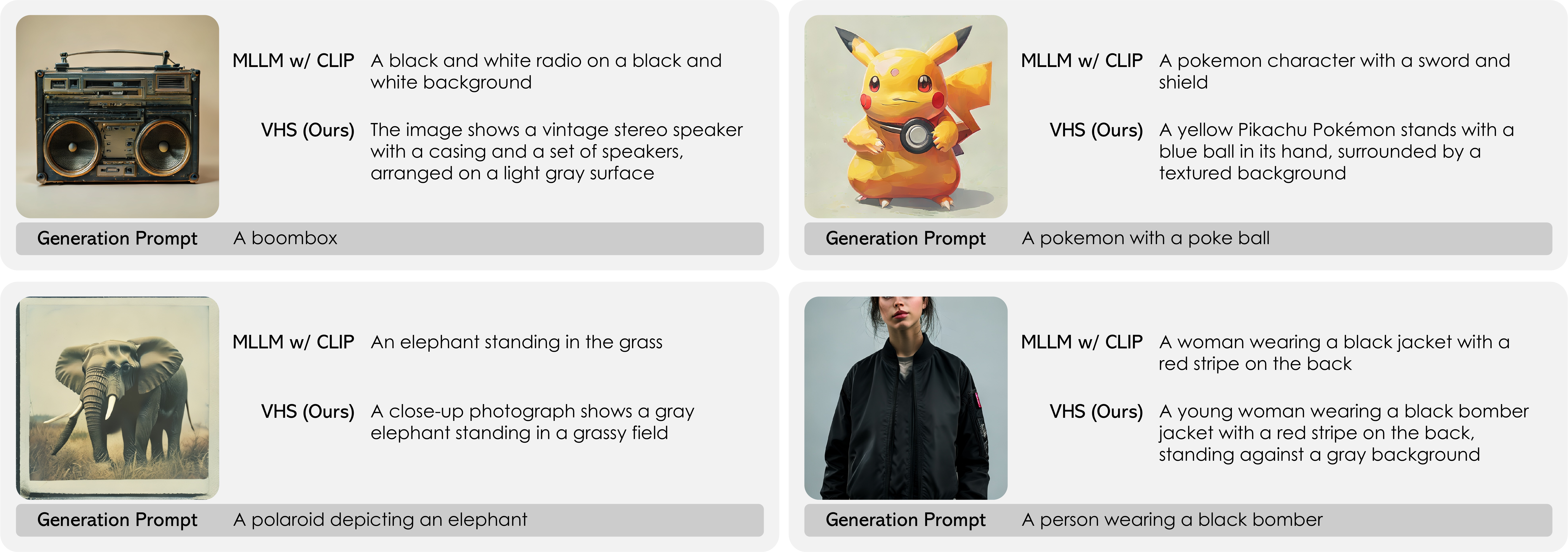}
    \vspace{-0.2cm}
    \caption{Qualitative results of captioning produced by \ours and MLLM w/ CLIP after the alignment training phase.}
    \label{fig:supplementary_stage_1_v1}
    \vspace{-0.2cm}
\end{figure*}

\begin{figure*}[t]
    \centering
    \includegraphics[width=0.98\linewidth]{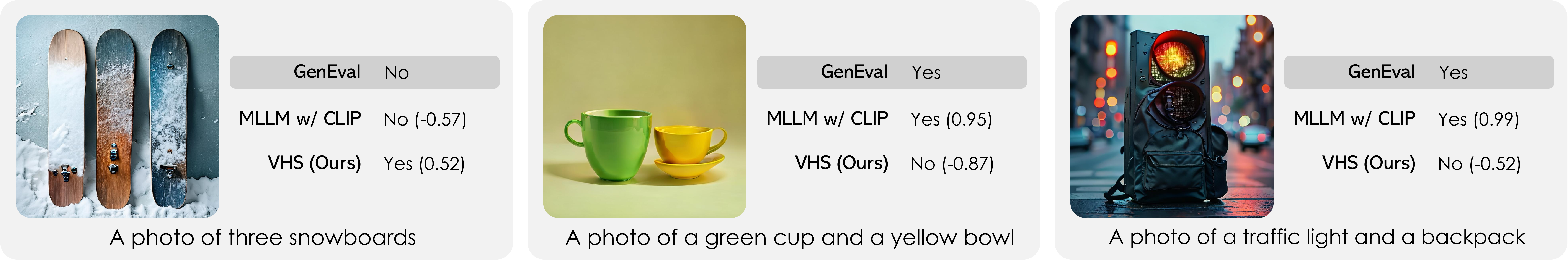}
    \vspace{-0.2cm}
    \caption{Cases where judgments from \ours is not aligned with the GenEval verifier.}
    \label{fig:supplementary_failure_cases}
    \vspace{-0.3cm}
\end{figure*}

\subsection{Robustness to Model Updates}
Acknowledging the tight coupling \ours introduces between the generator and its latent verifier, we investigate how distribution shifts in the generator's weights affect verification quality. To assess this, we fine-tune SANA-Sprint on an aesthetic dataset~\cite{wang2026moonworks} using LoRA~\cite{hulora} and measure GenEval performance within an 1100\,ms budget. Although distribution shift degrades overall performance, \ours exhibits significantly greater resilience than the pixel-space verifier (MLLM w/ CLIP), suggesting that the generator's hidden states remain semantically stable during fine-tuning. This relative stability is reflected throughout the training process. As observed in Table~\ref{tab:lora}, the performance gap between the methods grows from 1.4\% at the baseline to 1.8\% at mid-epoch, reaching 2.4\% after a full epoch.

\section{Additional Qualitative Results}
\label{supp:qualitatives}
\begin{figure*}[t]
    \centering
    \includegraphics[width=0.98\linewidth]{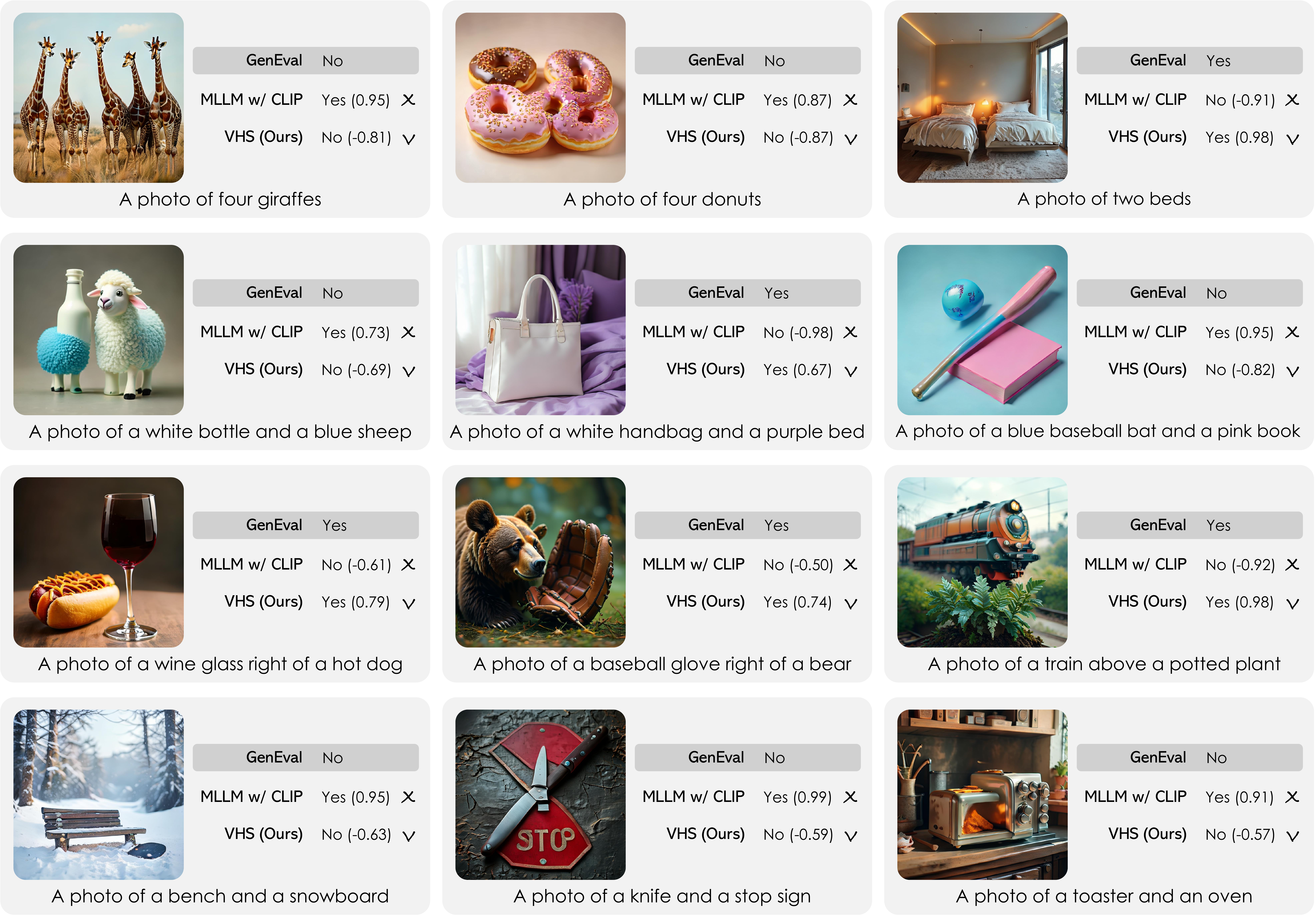}
    \vspace{-0.1cm}
    \caption{Visual comparison of the best pick images by different verifiers for images generated by SANA-Sprint~\cite{chen2025sana} on GenEval~\cite{ghosh2023geneval} prompts. $\checkmark$ and $\times$ express the alignment with the GenEval verifier.}
    \label{fig:qualitatives_supplementary}
    \vspace{-0.2cm}
\end{figure*}

In Fig.~\ref{fig:qualitatives_supplementary} we provide a qualitative evaluation of \ours using sample prompts from the GenEval benchmark with the SANA-Sprint generator, emphasizing its effectiveness in various challenging scenarios. \ours captures fine-grained details, distinguishing accurate images from those that are subtly incorrect. For instance, in the counting task involving ``four giraffes'', \ours correctly identifies the discrepancy in the image, which contains five giraffes, assigning a negative score consistent with the GenEval verifier. Similarly, \ours shows good performance in spatial reasoning, as it correctly validates the ``baseball glove right of a bear'' example which the baseline falsely rejects. These results further confirm that \ours accurately validates correct generations related to counting, attribute binding, and spatial relationships. 

Moreover, Fig.~\ref{fig:supplementary_stage_1_v1} presents qualitative examples of captions produced by \ours following only the alignment training phase. For each example, we generate sample images using the specified generation prompts and extract the corresponding hidden layer activations to feed into \ours. The examples show that \ours consistently produces accurate and detailed captions across diverse subjects. For instance, given the generation prompt ``A pokemon with a poke ball'', the generator produces an image which \ours then captions with a comprehensive description, identifying the character as ``a yellow Pikachu Pokémon'', and noting the ``textured background''. Similarly, when the generator creates an image from the prompt ``A boombox", \ours correctly describes it as ``a vintage stereo speaker with a casing and a set of speakers, arranged on a light gray surface'', capturing fine-grained visual details such as the surface color and arrangement.

\section{Limitations}
\label{supp:limitations}

While our proposed method demonstrates robust performance, we acknowledge certain limitations. Most notably, the verifier is coupled to the underlying generator architecture, as it operates directly on its hidden representations. This tight integration enables substantial efficiency gains and strong semantic alignment, but limits out-of-the-box transferability across substantially different generative backbones. In our experiments, we observe stable behavior under incremental model updates (Table~\ref{tab:lora}); however, if the architecture changes significantly, \ours must be retrained to maintain compatibility and verification performance. In practice, this constraint is of limited relevance in typical production settings, where generators are updated incrementally.

\begin{figure*}[t]
\centering
\begin{prompt}[1. Prompt for Tag Generation]
You are an assistant that classifies image generation prompts into one of six categories.
Given a text prompt describing an image, output ONLY the corresponding tag from this list:
- single_object
- two_object
- counting
- colors
- position
- color_attr

Rules:
1. Respond with exactly one of the tags above — nothing else.
2. Classification criteria:
   - "single_object" → only one object is mentioned.
   - "two_object" → exactly two different objects are mentioned (e.g. "a cat and a dog").
   - "counting" → a specific number of identical objects is requested (e.g. "three dogs").
   - "colors" → a single object with a color attribute (e.g. "a purple umbrella").
   - "position" → objects are described with a spatial relation (e.g. "a cat below a table", "a man left of a horse").
   - "color_attr" → two or more different objects, each with their own color (e.g. "a red apple and a green pear").
3. If uncertain, choose the most specific applicable tag.

Examples:
Input: "a photo of an umbrella" → single_object  
Input: "a photo of a bowl and a pizza" → two_object  
Input: "a photo of three persons" → counting  
Input: "a photo of a purple hair drier" → colors  
Input: "a photo of a couch below a cup" → position  
Input: "a photo of a red skis and a brown tie" → color_attr

The prompt is {input_prompt}.
\end{prompt}
\begin{prompt}[2. Prompt for Captioning on the Alignment Set]
Describe the image in one concise sentence. Be objective and precise, without speculation. Output only the description in plain text, without line breaks.
\end{prompt}
\begin{prompt}[3. Prompt for Image Scoring]
You are an AI assistant specializing in image analysis and ranking. Your task is to analyze and compare image based on how well they match the given prompt.  
<image> The given prompt is: {input_prompt}. 
Please consider the prompt and the image to make a decision and response directly with 'yes' or 'no'.
\end{prompt}
\vspace{-0.1cm}
\caption{Prompts employed for dataset generation and image scoring.}
 \label{fig:caption_prompt}
\end{figure*}

\end{document}